\documentclass{article}

\usepackage[dvipsnames]{xcolor}   
\usepackage[final]{neurips_2024}
\usepackage[utf8]{inputenc} 
\usepackage[T1]{fontenc}    
\usepackage{url}           
\usepackage{booktabs}       
\usepackage{amsfonts}      
\usepackage{nicefrac}       
\usepackage{microtype}    
\usepackage{graphicx} 
\usepackage[greek,english]{babel}
\usepackage{amsmath}
\usepackage{pifont}
\usepackage{adjustbox}
\usepackage{array}
\usepackage{makecell}
\usepackage{multirow}
\usepackage{tabularx}
\usepackage{booktabs}
\usepackage{wrapfig}
\usepackage{listings}
\usepackage{amssymb}
\usepackage{xspace}
\usepackage{authblk}
\usepackage{setspace}
\usepackage{appendix}
\usepackage{titletoc}
\usepackage{lscape}
\usepackage{pgf}
\usepackage{hyperref}
\usepackage{marvosym}
\usepackage{caption}
\usepackage{float}  
\usepackage{subfigure}  

\title{Richelieu: Self-Evolving LLM-Based Agents for AI Diplomacy}

\author{
Zhenyu Guan \textsuperscript{$\diamondsuit$}, Xiangyu Kong\textsuperscript{$\clubsuit \dagger$\Letter}, Fangwei Zhong\textsuperscript{$\spadesuit$$\dagger$\Letter}, Yizhou Wang\textsuperscript{$\heartsuit$$\diamondsuit$}\\
\textsuperscript{$\diamondsuit$} Institute for Artificial Intelligence, Peking University, Beijing, China\\
\textsuperscript{$\clubsuit$} Computer School, Beijing Information Science \& Technology University, Beijing, China \\
\textsuperscript{$\spadesuit$} School of Artificial Intelligence, Beijing Normal University, Beijing, China \\
\textsuperscript{$\heartsuit$} Center on Frontiers of Computing Studies, School of Computer Science,\\
                 Nat'l Eng. Research Center of Visual Technology, \\
                 State Key Lab of General Artificial Intelligence, Peking University, Beijing, China\\
\textsuperscript{$\dagger$} State Key Laboratory of General Artificial Intelligence, BIGAI, Beijing, China\\
 \Letter Corresponding authors: \texttt{xykong@bistu.edu.cn}, 
\texttt{fangweizhong@bnu.edu.cn} \\
 }

\begin{document}

\maketitle

\begin{abstract}
Diplomacy is one of the most sophisticated activities in human society, involving complex interactions among multiple parties that require skills in social reasoning, negotiation, and long-term strategic planning. 
Previous AI agents have demonstrated their ability to handle multi-step games and large action spaces in multi-agent tasks. However, diplomacy involves a staggering magnitude of decision spaces, especially considering the negotiation stage required. While recent agents based on large language models (LLMs) have shown potential in various applications, they still struggle with extended planning periods in complex multi-agent settings. Leveraging recent technologies for LLM-based agents, we aim to explore AI's potential to create a human-like agent capable of executing comprehensive multi-agent missions by integrating three fundamental capabilities: 1) strategic planning with memory and reflection; 2) goal-oriented negotiation with social reasoning; and 3) augmenting memory through self-play games for self-evolution without human in the loop.
\end{abstract}
\section{Introduction}

Diplomacy, a central element of international relations, is an intricate and multifaceted activity that lies at the heart of human society's most complex interactions.  It requires various skills such as social reasoning, negotiation, and long-term planning to manage relationships and alliances among multiple parties. Mirroring this complexity, the Diplomacy game~\citet{dip} involves seven players to control European powers, presenting a challenging strategic landscape that demands advanced negotiation and strategic planning to succeed.

The AI community has shown an increasing interest in the deployment of AI agents to master such games~\citet{Shoker2023ConfidenceBuildingMF, Konya2023DeliberativeTF, kramar2022negotiation, duenez2023social, mukobi2023assessing, kovač2023socialai}.
The recent breakthrough~\citet{Cicero} has turned into press diplomacy, which allows communication between players. 
However, the previous methods~\citet{Cicero} heavily rely on domain-specific human data, leading to its poor generalization to other scenarios/ applications.
The question then arises: \textbf{Can we build an AI agent that excels in the art of diplomacy without relying on domain-specific human data?}

Recently, agents based on the Large Language Model(LLM) have emerged as a promising development for AI agents. The previous applications on personal assistants~\citet{li2024personal}, robotics~\citet{cheng2024empowering, yang2023plug}, and video games~\citet{wan2024building} have shown the surprising ability of LLM-based agents in communication and planning, benefiting from the emergent ability of common sense reasoning, in-context/ few-shot learning, and sophisticated natural language processing on LLMs.
However, diplomacy presents a unique set of challenges. It not only requires planning long-horizon strategic~\citet{qi2024civrealm} and communicating with natural language, but also reasoning and adopting the complex social dynamics with partial observations, including gaining trust and reputation, building rapport, detecting deception, and assessing the reliability of other players.

In this work, we aim to make the first attempt to explore LLMs' potential to develop a human-like AI diplomacy agent. We name the agent Richelieu in memorizing a pivotal figure in European history who had enduring impacts on French politics, foreign affairs, and state building.
To achieve this goal, we have identified four core and essential capabilities that are crucial for building an LLM-based societal agent.
\begin{enumerate}
    \item {\bf Social reasoning.} This is the basic function for a social agent to interact with others, particularly for adapting to the dynamic changes in the nation's intentions and relationships.
    \item {\bf Balance long- and short-term planning.} Diplomacy necessitates a careful balance between short-term tactics and long-term strategies. An effective AI agent must assess the immediate consequences of its actions alongside their potential long-term impacts.

    \item {\bf Memory management.} A robust memory system is a critical component of learning and improvement. The AI agent must be able to recall and integrate information from past negotiations and actions to inform its current and future decision-making processes. This endows the agent with the ability to evolve.
    \item {\bf Self-reflection.}  An AI agent capable of profound reflection can analyze its own decisions, learn from its memory experience, and adapt its strategies accordingly. 
\end{enumerate}
By integrating these four capabilities, the agent can operate at the highest level of diplomatic sophistication, outperforming the state-of-the-art AI diplomats~\citet{Cicero}.

\begin{figure}[t]
    \centering
    \includegraphics[width=0.95\textwidth]{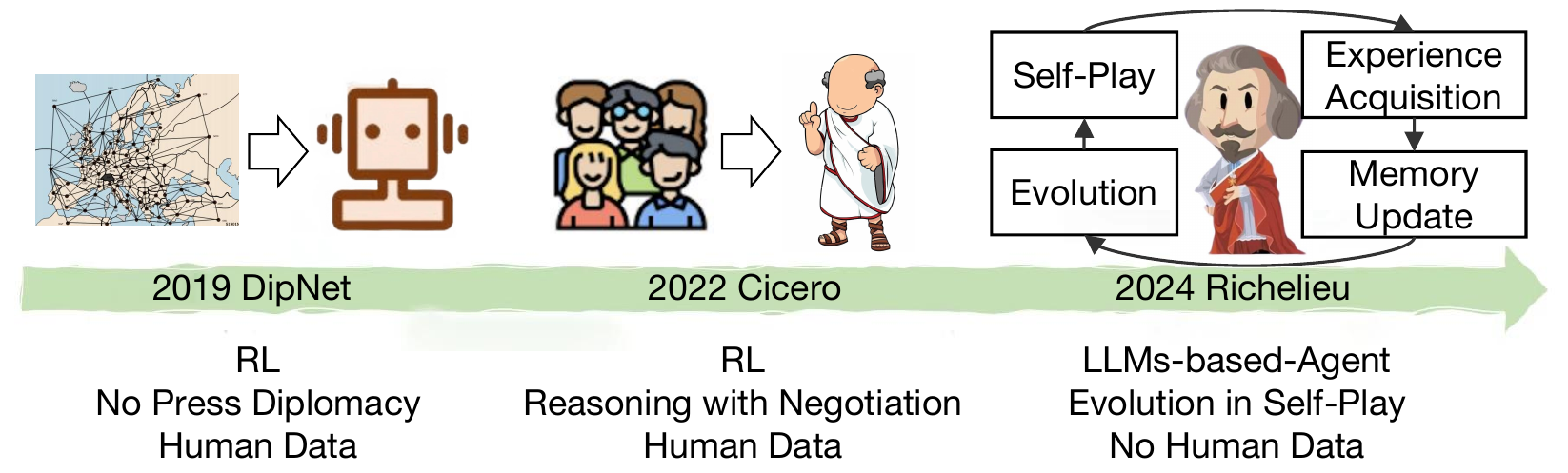}   
    \caption{A new paradigm for building AI Diplomacy agent.}
    \label{RvP}
\end{figure}

Our contributions can be summarized in three-fold: 1) We introduced a new paradigm for building AI diplomacy agents, compared to previous work (Fig.~\ref{RvP}). The agent can self-evolve by generating experience via self-play games, without any task-specific human data. 
2) We demonstrate the superior performance of our agent playing against the SOTA method, e.g., Cicero~\citet{Cicero}, that relies on a large-scale human demonstration for training. 
3) We further analyze the effectiveness of each module in our agent and the generalization of our agent in adopting different LLMs, such as \href{https://openai.com/index/gpt-4/}{GPT4.0} and
\href{https://llama.meta.com/llama3}{Llama 3}.

\section{Related work}
\label{gen_inst}

\textbf{AI Diplomacy.}
The game involves seven players controlling different powers in Europe. In each turn, players can negotiate for cooperation before making moves to take as many supply centers as they can. Apparently, this challenging strategy task requires both complex negotiation skills and superior planning capability for player agents to achieve final victory. So far, most previous works on this task remain focused on the planning strategies (a.k.a. {\bf No-Press Diplomacy} where no communication channels are allowed). The setting remains challenging considering its enormous action space of $10^21$ to $10^64$ per turn (compared with Chess, which has much fewer than 100 actions per turn). No wonder existing efforts rely on human data to play the game. Among the methods, one typical research is DipNet \citet{DipNet} which uses supervised and reinforcement learning. Based on DipNet, BRPI \citet{BRPI}, SearchBot \citet{SearchBot}, DORA \citet{DORA}, and KL-Regularized search (Diplodocus) \citet{jacob2022modeling} were conducted. Until very recently, research has also emerged for the full-setting of Diplomacy, or {\bf Press Diplomacy} where players are allowed to communicate with each other before making their moves in each turn. Such studies \citet{de2017d}\citet{Cicero}\citet{jaidka2024takes}\citet{kramar2022negotiation} mainly benefit from the recent thriving language models. Specifically, notable advancements include policy iteration methods from DeepMind and Facebook AI Research's equilibrium search agent \citet{jaidka2024takes}. However, Deepmind proposes to learn negotiation agents based on predefined contracts/protocols \citet{kramar2022negotiation}. And Meta AI's work, instead of one unified architecture, Cicero \citet{Cicero} integrates a language model for negotiation and an RL model for planning respectively. Such separately trained models make it inconvenient for agents' continual evolution. What's more, like no-press methods, these approaches heavily rely on human player data for agent training. Unlike these approaches, this paper delves into solving the negotiation and planning in one single self-evolving LLM-based agent model, without any pre-collected human expert training data.

\textbf{LLM-based Agents.}
With the emergence and growth of large language models (LLM), there is a growing trend in utilizing LLMs as fundamental controllers for autonomous agents\citet{wang2024survey}. One wide application genre is LLM-based answering engines, which merely cover the negotiation aspects of Diplomacy. Such systems include HuggingGPT~\citet{shen2023hugginggpt}, GPT4Tools~\citet{yang2023gpttools} and ToT~\citet{yao2023tree}, etc. They leverage LLMs to manage Al models, use tools, implement policy iteration, and enhance problem-solving across various tasks. Related work including AutoGPT, AgentGPT, BabyAGl \citet{Talebirad2023MultiAgentCH}, Toolformer \citet{schick2023toolformer}, and Visual ChatGPT aim to improve LLM capabilities in task automation and tool usage. Reflexion, a framework that improves LLMs through linguistic feedback and episodic memory \citet{zhang2024large}, facilitating better decision-making across diverse tasks is proposed. Besides 
\citet{NEURIPS2023_6b8dfb8c}\citet{wang2023voyager}\citet{Wang2023JARVIS1OM}\citet{zhu2023ghost}\citet{yan2023larp} apply LLM agents to the complex planning tasks in the well-known open-world game Minecraft\citet{fan2022minedojo}. Unlike these LLM-based agents which only focus on the negotiation/planning aspect, the proposed approach involves multiple self-evolving schemes to handle both of them simultaneously. 

\section{Problem Statement}
The Diplomacy game \citet{dip,inv} is set in pre-World War I Europe and involves each player (agent) representing one of the seven Great Powers of Europe, such as Germany, France, England, Italy, Austria-Hungary, Russia, and Turkey. Each player has a set of military units, including armies and fleets, which they can move and use to capture other supply centers.
The ultimate goal for the agent is to control a majority of the total supply centers on the board by the end of the game's Fall phase.
It's important to note that it is not won by eliminating other players or their units; it is won by controlling the requisite number of supply centers. This often involves forming and breaking alliances, negotiating, and sometimes betraying other players to achieve one's own goals. 

In each turn, the agent $i$ gets the current state $s_t\in S$, the actions of other players from the previous turn $\vec{a}^{-i}_{t-1}$, and the messages $\vec{m}^{-i, i}_{t}$ from other players during this turn's negotiations. The state $s_t$ for the environment includes the ownership of each territory on the map by a particular country and where the armies of each country are located.
Based on this information, the agent needs to engage in negotiations with other players, sending messages $\vec{m}^{i, -i}_{t}$ to chat with other players, and then take the actions $a^i_t$ in this turn.
The possible actions an agent can take $a^i_t \in A$ are commands to the armies, such as moving into an adjacent territory, supporting another unit, or holding a position.  Actions can also include diplomatic moves, such as proposing or withdrawing from an alliance, although these are less formalized in the game mechanics.\citet{DipNet, rul} 

\section{Self-Evolving LLM-based Diplomat}

\begin{figure}[t]
    \centering
    \includegraphics[width=0.95\textwidth] {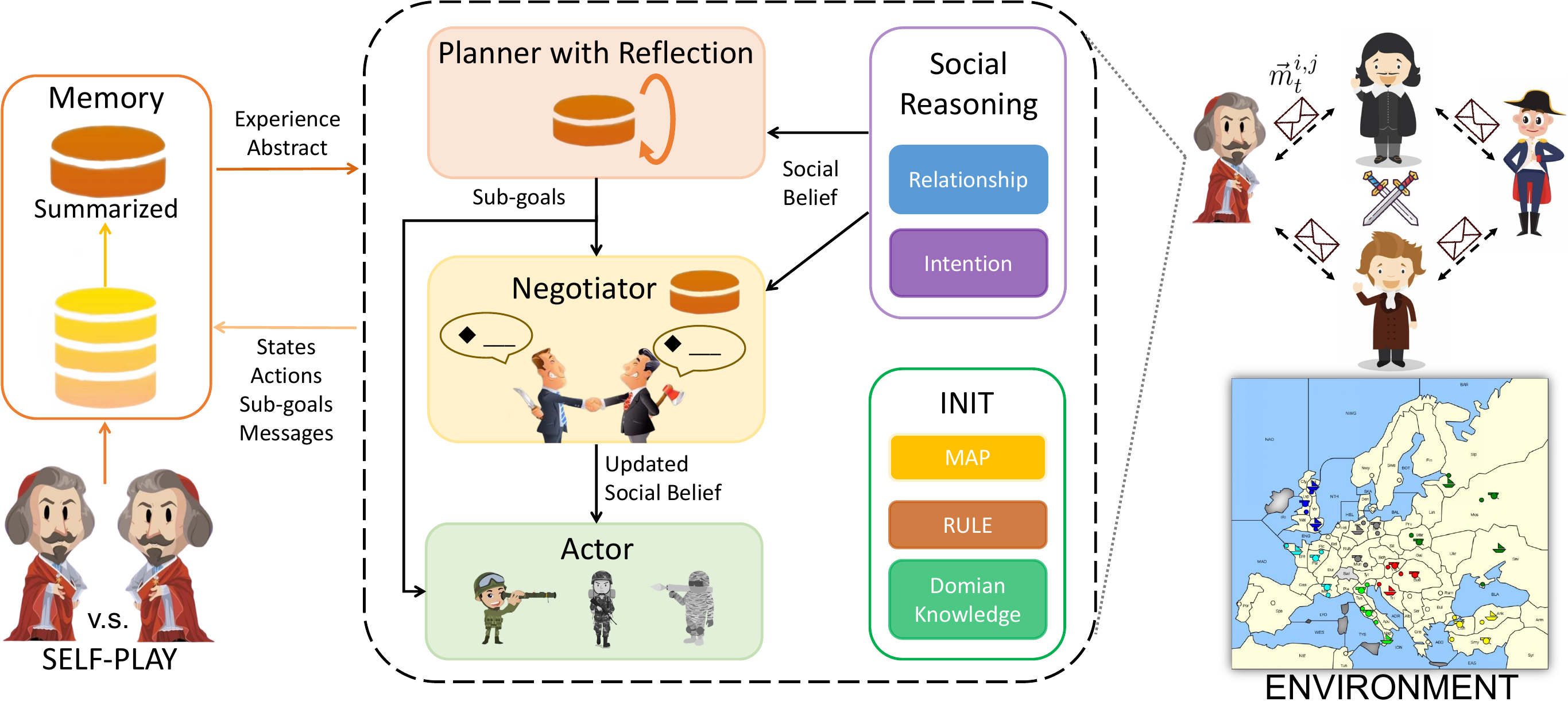} 
    \caption{The framework of the proposed LLM-based-agent, Richelieu. It can explicitly reason social beliefs, propose sub-goals with reflection, negotiate with others, and take actions to master diplomacy. It augments memories by self-play games for self-evolving without any human annotation.
    }
    \vspace{-0.3cm}
    \label{richelieu}
\end{figure}

We have constructed a comprehensive framework with modules for memory management, social reasoning, strategic planning, negotiation, decision-making, memory update, and self-evolving to fully leverage the capabilities of LLMs. Richelieu starts by setting up with map details, game rules, domain knowledge, and the long-term goal.\citet{zhang2022automatic,wei2022chain,wang2022self} At each turn, the agent will run in the following steps:
1) \textbf{Social Reasoning:} First of all, the agent undergoes a comprehensive analysis of the game state $s_t$ to build the social belief, including the intention of other players and their relationship $\vec{\phi_{t}} \in \Phi^{n}$.\citet{zhang2024llm,gurcan2024llm}
2) \textbf{Planner with Reflection:} Then, the agent proposes sub-goals $\chi^i_{t} \in X$ that is strategically aligned with the long-term goals $\Upsilon$, with the social belief and refining the proposed goal with experience $\vec{\eta_t}\in H^m$ abstract from the memory $M$ via self-reflection.\citet{wang2024devil, wang2024re2llm}
3) \textbf{Negotiator:} To achieve the sub-goals, the negotiator will start a dialogue session with some players, and evaluate their trueness $\vec{\psi}^{-i}_{t}$  by referring to their words $\vec{m}^{-i, i}_{t}$, the current state $s_t$, their sincerity $\vec{\gamma}^{-i}_t$ and the experience $\vec{\xi_t}$ .\citet{abdelnabi2023llm,bianchi2024well}
4) \textbf{Actor:} After negotiation, the actor decides its course of action $a^i_{t}$, based on the sub-goal $\chi^i_{t}$ and updated social state $s_{t+1}$, marking the end of that turn. 
5) \textbf{Memory Management:} The state of the current turn $s_t$, the content of negotiations $\vec{m_t}$, the actions taken by all players $\vec{a_t} \in A^n$, and the sub-goals set forth $\chi^i_{t}$ are all logged within the memory as $\mu \in M$. This logged data serves as a historical experience, guiding Richelieu's subsequent actions in future turns~\citet{hatalis2023memory, zhang2024survey}.
6) \textbf{Self-evolution:} The agent's evolution is highly dependent on the diversity of experiences stored in its memory.
As this diversity grows, so does the agent's capability.
Without human demonstrations, we employ multi-agent self-play games, i.e., our agents respectively control all the countries to simulate and acquire diverse experiences for self-evolving. Notably, the agent can further evolve during testing to adapt to different players.

\subsection{Social Reasoning}
There are no permanent enemies, no permanent allies.
The relationship among countries is dynamically changing upon the evolving global state.
However, it is difficult to determine the appropriate allies and enemies with partial observation. For example, there is uncertainty about the intentions of potential allies, which could lead to betrayal at pivotal moments. Consequently, we need to identify the intention and relationship of the current state by social reasoning to shape the social belief \citet{zhang2024llm,gurcan2024llm}. 

\textbf{1) Modeling Relationship:} 
Before setting sub-goals, Richelieu evaluates its relations with others, identifying enemies such as aggressive nations, vulnerable neighbors for expansion, and those with long-term potential threats. It also seeks out potential allies to counter these threats.\citet{sun2024llm, zhang2024towards}
Simultaneously, Richelieu also tries to identify potential allies that could be instrumental in countering these adversaries. By isolating the analysis of inter-player relationships as a discrete element, Richelieu strategically exploits the actions of other players in subsequent stages of the game to reach its goals. 
\textbf{2) Inferring Intention:}
The social belief is used by the planner, ensuring that its sub-goals are formulated with a comprehensive consideration of the behaviors and intentions of other intelligent agents within the game. Richelieu's sub-goals will particularly emphasize on those who are identified as potential adversaries or allies, fostering more effective collaboration with potential allies and participation in strategic opposition against adversaries. Furthermore, the insights gleaned from this analysis are instrumental in the subsequent negotiation phases. They are employed to assess the authenticity of the statements made by other players, as well as to aid Richelieu in reaching cooperative agreements.\citet{de2023emergent,he2024llm}

\subsection{Strategic Planner with Reflection}

The strategic planner specifies the sub-goals, which serves as an intermediary between immediate actions and the overarching goal of securing victory in the game.  
That is because we observe that LLMs are often characterized by their propensity to prioritize short-term gains in decision-making processes, with a notable deficiency in incorporating the future into their strategic calculations. \citet{renze2024self,zhang2024agent}For example, it is common for a non-neighboring country to become too powerful.
Formally, $\vec{\chi_t} \gets SR(s_t, \vec{\phi_t}, \Upsilon)$
where $\vec{\chi_t}=(\chi^i_{t},\chi^1_{t},\ldots,\chi^n_{t})$ represents the proposed sub-goals and other players' intention that we inferred, $ \vec{\phi_t} \in \Phi^n$ represents the inferred relationship on the social belief.
These goals may encompass a range of tactical considerations, such as the containment of a formidable rival's advancement or the strategic expansion in a particular direction to consolidate power.

\textbf{Reflection with Memory.} We further develop a reflection mechanism to enhance the rationality and effectiveness of our agent's sub-goals in achieving long-term goals.\citet{liu2024agentlite} This reflection mechanism relies on the past experiences to critique and enhance proposed sub-goals. 
We employ a similarity-based function to find relevant historical experiences that match the current game state from its memory. This function considers two factors: goal similarity and state similarity, to select the most comparable experiences. The process can be written as:
$\vec{\eta_t}\gets h(s_t,\chi^i_{t},M)$, where $\vec{\eta_t}\in H^m$. In practice, considering the limited context windows of LLM, we retrieve the most analogous experiences from the memory based on these metrics.
Experiences with high evaluative scores reinforce successful strategies and support the continuity of existing sub-goals. On the other hand, lower scores indicate areas that need improvement and prompt the necessary adjustments.
As our agent, Richelieu, undergoes more training sessions, its reflection abilities improve. The growing pool of historical experiences consistently enhances its performance.

\subsection{Negotiator and Actor}
By chatting with other players, the goal of the negotiation is to update the social belief according to the received words and reach the sub-goal by manipulating other's intentions, such as securing cooperative agreements with other nations, terminating ongoing conflicts with a specific country, or deterring the formation of alliances directed against its interests.\citet{noh2024llms,zhan2024let} However, it is difficult to reach a consensus, as the interests and strategies of the various nations often conflict, and trust between players can be scarce, making it challenging to establish and maintain cooperative agreements.
In this case, we argue that the negotiator should identify the true intentions and relationship of the opponent before generating the words for the negotiation. 

\begin{figure}[t]
    \centering
    \includegraphics[width=0.9\textwidth]{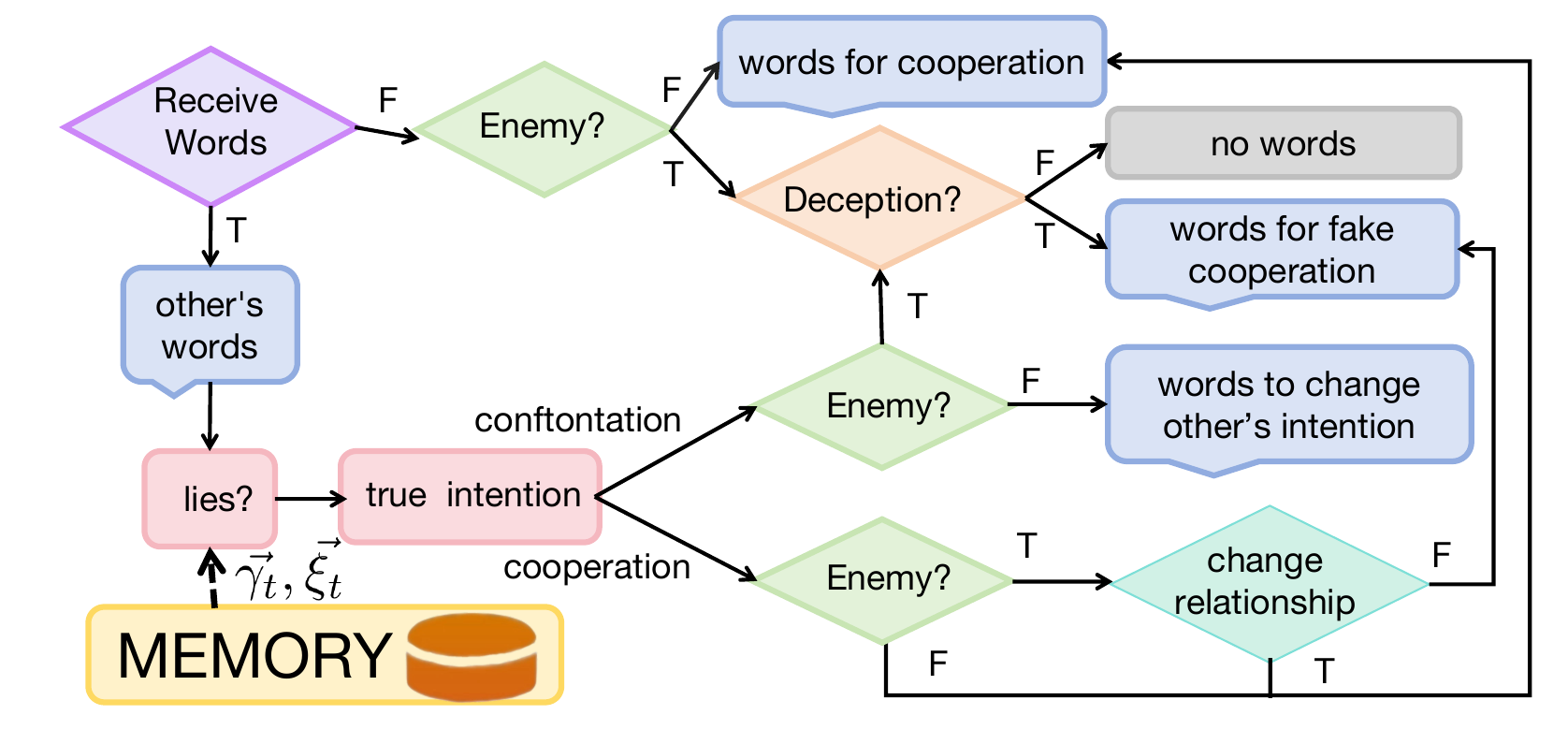} 
    \vspace{-0.1cm}
    \caption{The social reasoning flow for negotiation. With the received words and memory, the agent will reason by answering the following questions: ``Is the opponent lying?", ``What is the true intention of the opponent?", ``is the opponent enemy?", ``Is it necessary to deceive the opponent?", and ``Is it necessary to change the relationship with the opponent?", and then generate the words accordingly for negotiation.}
    \label{negotiation}
    \vspace{-0.3cm}
\end{figure}

To fully utilize the power of LLMs, we construct a social reasoning flow for negotiation, as shown in Figure~\ref{negotiation}. During the negotiation process, we guide Richelieu to consider the veracity of what other players said and their true intentions, and in conjunction with our established sub-goals and analysis of our relationships with other players, to negotiate and form alliances with potential allies and attempt to deceive enemies.\citet{xia2024measuring,moghimifar2024modelling}

To counteract the challenge of non-binding agreements and potential deception, we incorporate a discrete module dedicated to the assessment of the veracity of statements made by other players during negotiations. To determine the truthiness of other players' statements $\psi^j_{t}$, three main factors are considered. The most important is the consistency between the player's sub-goals $\chi^{j}_{t}$ that our agent inferred before and the intentions conveyed through his statements $m^{j,i}_{t}$. To aid in the judgment, our agent also goes through the memory to retrieve the consistent experiences $\vec{\xi_{t}}$. Additionally, the player's overall honesty score $\gamma_i$ is taken into account. Hence, we get the truthiness of the opponent $j$: $\psi^j_{t} \gets g(s_t, \chi^j_{t}, m^{j,i}_{t}, \vec{\phi_t}, \gamma_j, \vec{\xi_{t}})$, where $\vec{\xi_{t}}=w(s_t,m^{j,i}_{t},M)$.
With such a reasoning flow, our agent can adeptly navigate diplomatic discourse. 
After the negotiation, the actor will get the updated social beliefs and choose a specific action for the army.

\subsection{Memory Management and Evolution in Self-Play Games}
\label{sec:mem}
This memory is the foundation of the framework that accumulates the historical experience of the agent and summarizes them for other modules.\citet{gao2024memory,li2024hello,yu2024finmem,hou2024my}
It supports other modules, such as planner and negotiator, to provide long-tail experiences.

\textbf{Raw Experience Management.}
Specifically, the memory module is tasked with the acquisition and archival of historical data, encompassing the observed game state $s_t$ at each turn, its sub-goals $\chi^i_{t}$, the messages during the negotiation $\vec{m_t}$, and the actions of all the players $\vec{a_t}$. Subsequently, the raw experience is summarized in a shorter content with an evaluation $\lambda_t \in \Lambda$ of the proposed sub-goals and an assessment of the credibility of other players $\gamma_j \in \Gamma$. 
$\lambda_t$ serves to reflect upon the agent's sub-goals. It evaluates whether sub-goals are reasonable based on the subsequent state and long-term goals $\Upsilon$. As the game progresses, it is continuously updated in response to changes in the state $\lambda_t \gets f(\chi^i_{t}, \Upsilon, \vec{s})$, where $\vec{s}=(s_t,s_{t+1},\ldots s_T) $. The formula represents the update of the evaluation $\lambda_t$ for the sub-goal in turn $t$  by the memory in turn $T$.
The updates will cease when there is a fundamental change in the sub-goal compared to the goal at turn $t$. This prevents subsequent decisions from impacting the assessment of the current decision-making.
We employ $\gamma_j \in \Gamma$ to evaluate the credibility of player $j$ and utilize $\tau^j_{t} \in \{0,1\}$ to denote the truthfulness, i.e., whether the statements made by the player $j$ during the negotiation process at time $t$ are truthful. The truthiness of player $j$'s statements is updated according to the memory from the previous turns, $\tau^j_{t}\gets T(s_t,s_{t+1}, a^j_{t}, m^{j,i}_{t})$.
The credibility of player $j$ $\gamma_j$ will be updated based on player $j$'s statements $\tau^j_{t}$, written as $\gamma_j \gets p(\gamma_j,\tau^j_{t-1})$.
Players' credibility $\vec{\gamma}$ is a short-term memory that is applicable only to the current turn. Other data collected or generated constitutes long-term memory. These data will be combined to form a history $\mu \in M$, and then is incorporated into memory.

\textbf{Acquisition Experience via Self-Play Games.} Self-play allows the agent to accumulate more experiences for self-evolution.\citet{liu2024agentlite,zhang2024large} After training, when Richelieu is faced with a certain state, it can draw on a larger pool of similar historical experiences. Diverse evaluations enable Richelieu to reflect more comprehensively on the strategies it currently devises, leading to a stronger optimization of decision making. 
As self-play continues, the acquisition of new and better historical experiences by Richelieu will diminish. This means that Richelieu's capabilities will not improve indefinitely. At the same time, as the memory grows, selecting appropriate historical experiences becomes a new challenge. The chosen m experience $\vec{\eta_t}$ may be almost identical, which could actually reduce the amount of useful information available to Richelieu.
As shown in Figure~\ref{train}, Richelieu's performance against Cicero \citet{Cicero} becomes better with increasing training iterations. With the accumulation of experiences, Richelieu's win rate exhibited a steady increase with accumulated training iterations, ultimately plateauing at a stable performance level. In contrast, the defeated rate showed a consistent decrease, approaching an asymptotic value. These observations confirm the effectiveness of self-play in Richelieu's evolution.

\section{Experiment}
In the experiments, our goal is to answer the following questions: 
1) \textbf{Mastery of Non-Press Diplomacy}: Can our agent master the non-press diplomacy against baselines? 
2) \textbf{Competing with State-of-the-Art}: Can our agent surpass the performance of the current state-of-the-art agents in press diplomacy? 
3) \textbf{Compatibility with LLMs}: Can our self-evolving framework be compatible with different LLMs?
4) \textbf{Contribution of Framework Modules}: Do the individual modules within our framework contribute to the overall improvement of our agent's performance? 
5) \textbf{Social Reasoning}: Can Richelieu accurately infer the true intentions of other players and reasonably determine the relationships of ally or enemy with them? 
The implementation of our method can be found at:
\url{https://github.com/todexter3/Richelieu.git}

\subsection{Experimental Setup}
\textbf{Environment.} The widely-used open source Diplomacy game platform introduced by \citet{DipNet} is adopted for evaluating Richelieu against other models. It is easy to switch between no-press (with negotiation between players) and press (no negotiation between players) games based on this platform, facilitating comparison on both settings. The platform also contains over 10,000 human game data on which previous approaches are trained. Note that our method does not need them. In each game, a model will play the role of one randomly selected country to compete against countries controlled by other methods. It wins if occupying all the supply centers and loses vice versa.

\begin{table}[t] 
    \centering
    \caption{The results of our method playing against Cicero.}
    \label{against Cicero}
    \begin{minipage}{0.49\textwidth}
        \centering
        \adjustbox{max width=\textwidth}{
        \begin{tabular}{lllll}

            \toprule
            Model  & Win\textcolor{blue}{$\uparrow$}  & Most SC\textcolor{blue}{$\uparrow$}  & Survived\textcolor{blue}{$\uparrow$}  & Defeated\textcolor{orange}{$\downarrow$}\\
            \midrule
            Richelieu\_1&6.20\%	&9.40\%	&38.90\%	&45.50\%\\
            Richelieu\_2&6.60\%	&7.80\%	&40.80\%	&44.80\%\\
            Richelieu\_3&7.10\%	&9.30\%	&39.90\%	&43.70\% \\
            Richelieu\_4&7.40\%	&8.00\%	&40.20\%	&44.40\%  \\
            \midrule
            Cicero\_1&5.90\%	&6.50\%	&41.50\%	&46.10\%\\
            Cicero\_2&6.30\%	&7.20\%	&42.50\%	&44.00\%  \\
            Cicero\_3&5.90\%	&7.00\%	&41.60\%	&45.50\%  \\
            \midrule
            \textbf{Richelieu}& \textbf{6.83}\%	&\textbf{8.63}\%	&\textbf{39.95}\%	&\textbf{44.60}\%\\
            \textbf{Cicero}&\textbf{6.03\%}	&\textbf{6.90\%}	&\textbf{41.87\%}	&\textbf{45.20\%}\\
            \bottomrule
        \end{tabular}
        }
    \end{minipage}
    \hfill  
    \begin{minipage}{0.49\textwidth}
        \centering
        \adjustbox{max width=\textwidth}{
        \begin{tabular}{lllll}
            \toprule
            Model  & Win\textcolor{blue}{$\uparrow$}  & Most SC\textcolor{blue}{$\uparrow$}  & Survived\textcolor{blue}{$\uparrow$}  & Defeated\textcolor{orange}{$\downarrow$}  \\
            \midrule
            Richelieu\_1&6.30\%	&7.90\%	&39.40\%	&46.40\%\\
            Richelieu\_2& 6.60\%	&8.30\%	&41.20\%	&43.90\%\\
            Richelieu\_3& 7.20\%	&8.70\%	&41.70\%	&42.40\% \\
            \midrule
            Cicero\_1&5.80\%	&6.70\%	&41.20\%	&46.30\%\\
            Cicero\_2 &6.50\%	&7.20\%	&42.50\%	&43.80\%\\
            Cicero\_3&6.00\%	&7.00\%&41.60\%	&45.40\%\\
            Cicero\_4&6.10\%	&7.20\%	&42.30\%	&44.40\%\\
            \midrule
            \textbf{Richelieu} &\textbf{6.70\%}	&\textbf{8.30\%}	&\textbf{40.77\%}	&\textbf{44.23\%}\\
            \textbf{Cicero}&\textbf{6.10\%}	&\textbf{7.03\%}	&\textbf{41.90\%}	&\textbf{44.98\%}\\
            \bottomrule
        \end{tabular}
        }
    \end{minipage}

\end{table}

\textbf{Evaluation Metrics.}
We evaluate the models based on the results of multiple rounds of games. In each round, the model is randomly assigned a country to control. Typically, 1000 rounds are played to obtain the average results. We evaluate the models in two metrics. One is based on the win rate, Most SC rate, survived rate, and defeated rate. There are four possible outcomes for each country in the game. If a country loses all its supply centers (SC), it is eliminated and recorded as ``defeated". If a country occupies 18 or more out of 34 supply centers, the game ends, and that country is recorded as ``win", while other countries are recorded as ``defeated". In other cases, the game ends in a draw. The country with the most supply centers is recorded as ``Most SC", the countries that have been eliminated are recorded as ``defeated", and the other countries are recorded as ``Survived".
The other is based on the scores obtained by the models after multiple rounds of competition. To compare the capabilities of multiple models, we use C-Diplo Argir\citet{GiraudonArcher2024}, a scoring system. This system is used in many international diplomacy competitions. The scoring method is as follows: If a player wins by occupying 18 or more supply centers, the player scores 93 points, and each of the other six players scores 1 point. If the game ends in a draw, the player with the most centers scores 37 points. The second player with the most centers scores 14 points. The third player with the most centers scores 7 points. Each player scores 1 point per center owned. Each player also scores 1 point for participating. In this way, regardless of the game outcome, a total of 99 points will be distributed among the players in each game.

\textbf{Baselines.} We select six previous models as baselines for comparison. Among them, Cicero\citet{Cicero} by Meta is a diplomacy model with a negotiation module. The SL-DipNet and RL-DipNet \citet{DipNet}, the BRPI \citet{BRPI}, the SearchBot \citet{SearchBot}, and the DORA\citet{DORA} are no-press diplomacy models. We also build a LLM-based agent, AutoGPT \citet{yang2023auto}. In experiments, we set a temperature of 0.3 to ensure a relatively stable generation of LLM policies. The overall reasoning framework also ensure the stability and consistency in the AI agent's performance. 

\subsection{Results}

\begin{wrapfigure}[20]{r}{0.5\linewidth}
    \centering
    \vspace{-0.4cm}
    \includegraphics[width=\linewidth]{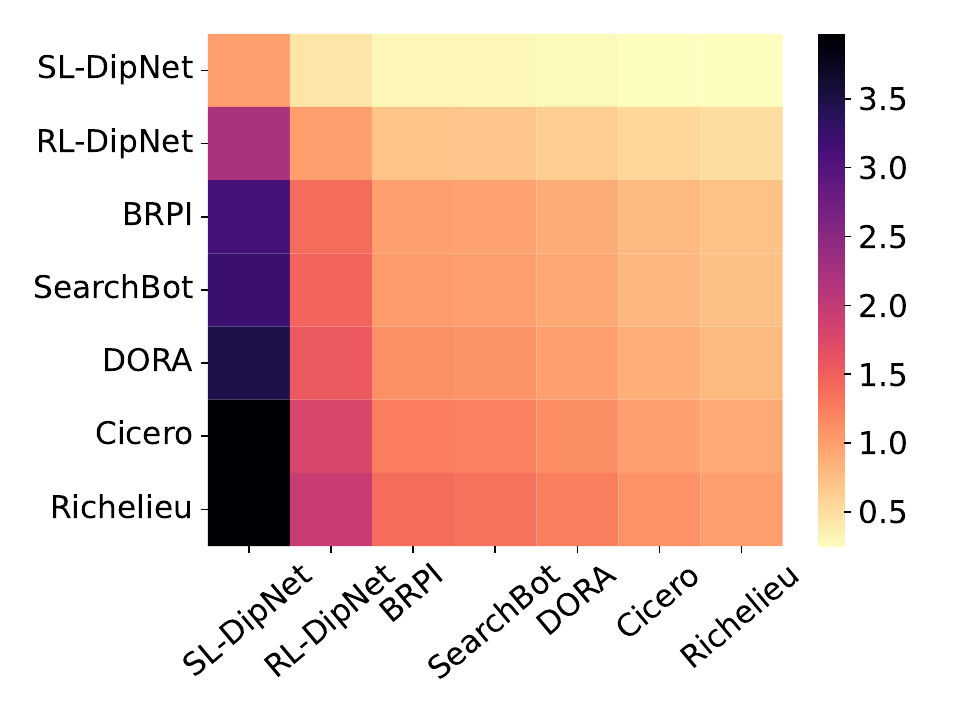} 
    \vspace{-0.5cm}
    \caption{The relative scores among 7 different agents when massively playing on the no-press setting. Each point shows the ratio of the model's score on the vertical axis to the score gained by the model on the horizontal axis.}
    \label{massive}
\end{wrapfigure}

\textbf{Massively Play with Baselines on no-press setting.} We let Richelieu compete with the other six models including Cicero\citet{Cicero}, SL-DipNet and RL-DipNet \citet{DipNet}, BRPI \citet{BRPI}, SearchBot \citet{SearchBot}, and DORA\citet{DORA} on No-Press Diplomacy, in which players make moves without communication. Figure ~\ref{massive} indicates that Richelieu outperforms other previous models relying on human game data. In contrast, Richelieu does not need such data but outperforms these methods by a clear margin, which demonstrates the outstanding planning capability of Richelieu.

\textbf{Play against Cicero on press setting.} We also evaluate Richelieu through competition against Cicero in the challenging scenario where negotiation is enabled. Specifically, we randomly assign three countries to one model and the remaining four to another. After playing several rounds of the game, the win rate, most SC rate, survived rate, and the defeated rate is calculated using a weighted average for evaluation. Table ~\ref{against Cicero} demonstrates the competitive performance of Richelieu in comparison to Cicero. Richelieu's win rate is approximately 0.7\% higher than Cicero's. If the Most SC rate is also taken into account, Richelieu is about 2\% higher than Cicero. At the same time, Richelieu's loss rate is also 0.6\% lower. According to our scoring system, Richelieu's score is about 10\% higher than Cicero's. This is nontrivial especially when Richelieu is trained in a self-play game without humans and the opponents are trained with the data from human players. 

Although Richelieu's win rate improvement compared to Cicero is not significant, the relative value of the improvement is quite large. Moreover, the main reason for the modest improvement is that in the seven countries, there are three or four controled by Richelieu with similar abilities, which often results in the game ending in a draw. Moreover, we observed a large gap by comparing the scores the agents gained in the massively play with baselines on no-press setting showed in Figure~\ref{massive}. Our agent's score is about 10\% higher than Cicero's.

\begin{wraptable}[11]{r}{0.6\linewidth}
\small
\vspace{-0.4cm}
  \caption{The results of our method playing against AutoGPT.}
  \label{llmbase}
  \centering
  \begin{tabular}{lrrrr}
    \toprule
    Model  & Win\textcolor{blue}{$\uparrow$}  & Most SC\textcolor{blue}{$\uparrow$}  & Survived\textcolor{blue}{$\uparrow$}  & Defeated\textcolor{orange}{$\downarrow$}\\
    \midrule
     Richelieu\_1&9.30\%	&18.20\%	&37.90\%	&34.60\%\\
    Richelieu\_2&9.90\%	&19.40\%	&37.70\%	&33.00\%  \\
    Richelieu\_3&8.10\%	&17.40\%	&39.20\%	&35.30\% \\
    \midrule
    AutoGPT\_1&1.20\%	&4.60\%	&32.40\%	&61.80\%\\
    AutoGPT\_2&1.20\%	&4.20\%	&34.40\%	&60.20\% \\
    AutoGPT\_3&1.50\%	&4.00\%	&32.50\%	&62.00\% \\
    AutoGPT\_4&2.60\%	&3.60\%	&32.30\%	&61.50\% \\
    \midrule
    \textbf{Richelieu}& \textbf{9.10\%}	&\textbf{18.33\%}	&\textbf{38.27\%}	&\textbf{34.30\%}\\
    \textbf{AutoGPT}&\textbf{1.63\%}	&\textbf{4.10\%	}&\textbf{32.90\%}	&\textbf{61.37\%}\\
    \bottomrule
  \end{tabular}
\end{wraptable}

\textbf{Play against AutoGPT on press setting.}
We further built an LLM-based agent using AutoGPT and compared it with our agent. In the testing, we randomly select three countries to be controlled by Richelieu, and the other four countries to be controlled by AutoGPT. Note that the agent controls each country independently. The results are showed in Table ~\ref{llmbase}. The results show that our model outperforms the existing LLM baseline.

\begin{figure}[t]
    \centering
    \includegraphics[width=1.0\textwidth]{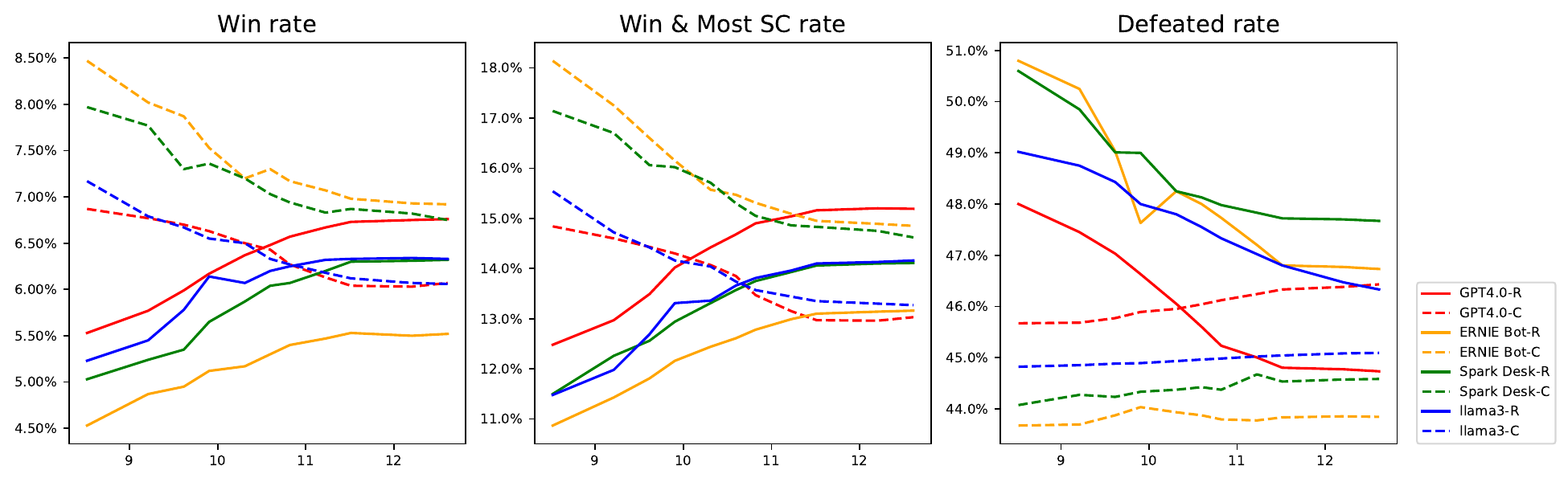} 
    \caption{Richelieu modules benefit different LLMs. The solid line represents the experimental results for Richelieu, while the dashed line corresponds to Cicero. Different colors are used for different LLMs. The horizontal axis represents the logarithm of the number of training sessions, and the vertical axis denotes the rate.}
    \label{train}
\end{figure}

\newpage
\textbf{Generalization of self-evolving framework to different LLMs.}
To demonstrate the effectiveness of our framework in a variety of LLM, we conducted experiments using four models: \href{https://openai.com/index/gpt-4/}{GPT4.0}, \href{https://yiyan.baidu.com/welcome}{ERNIE Bot}, \href{https://xinghuo.xfyun.cn/}{Spark Desk}, and \href{https://llama.meta.com/llama3}{Llama 3}. As the number of training iterations increases, Richelieu's win rate steadily improves while the defeated rate declines, ultimately reaching a relatively stable outcome. This suggests that our self-play method is effective. After training, the win rate using GPT4.0 increased from 1.5\% lower than Cicero's to about 0.7\% higher than Cicero's. The win rate using llama3 increased from 2.3\% lower than Cicero's to almost equal to Cicero's. The win rates using Models Spark Desk and ERNIE Bot increased from 3\% and 4\% lower than Cicero's to 0.7\% and 1.6\% lower than Cicero's, respectively. The experimental results show that, despite variations in Richelieu's performance due to the inherent differences in the capabilities of these LLMs, as illustrated in Figure~\ref{train}, our framework and training approach significantly enhance the capabilities of all LLMs.This indicates the generalization of a self-evolving framework to various LLMs.

In order to demonstrate the effect of the memory from the self-play game on the strategy of our agent, we found two turns with similar states in different rounds, one before self-play and the other after. The cases are showed in Appendix~\ref{cs}. 
\begin{table}[t]
  \caption{Ablation study: average results of 3 Richelieu vs. 4 Cicero.}
  \label{development}
  \centering
  \begin{adjustbox}{width=1.0\textwidth}
  \begin{tabular}{cccccllll}
    \toprule
    \makecell{Modeling\\ others}& \makecell{sub-goals}& \makecell{Negotiation \\pipeline} & \makecell{Reflection\\with Memory}&Self-play&Win \textcolor{blue}{$\uparrow$} &Most SC\textcolor{blue}{$\uparrow$}&Survived\textcolor{blue}{$\uparrow$}&Defeated\textcolor{orange}{$\downarrow$}\\
    \midrule
     & & & & &0.4\%&0.7\%&4.3\%&94.6\%\\
    \midrule
    \ding{51}&&&&&0.7\%&1.2\%&10.6\%&87.5\%\\
    \midrule
    \ding{51}&\ding{51} & &&&3.3\%&4.7\%&26.7\%&65.3\%\\
    \midrule
    \ding{51}& \ding{51} & \ding{51} & &&3.8\%&5.8\%&33.1\%&57.3\%\\
    \midrule
    \ding{51}&\ding{51}&\ding{51}&\ding{51}&&5.2\%&6.6\%&39.5\%&48.7\%\\
     \midrule  \ding{51}&\ding{51}&\ding{51}&\ding{51}&\ding{51}&\textbf{6.7}\%&\textbf{8.5}\%&\textbf{40.4}\%&\textbf{44.4}\%\\
   
    \bottomrule
  \end{tabular}
  \end{adjustbox}
\end{table}

\textbf{Ablation Study.}
We conduct comprehensive ablation studies on Richelieu by analyzing the benefit of incorporating Richelieu's various modules, like planners or memory, into basic LLMs. The results are shown in Table~\ref{development}. As illustrated in Figure ~\ref{train}, while the enhanced alignment in LLMs indeed boosts performance (GPT-4.0 is better than others), we observed that a vanilla GPT-4.0 still falls short in AI diplomacy without our framework, as can be seen in Table ~\ref{development}. Richelieu's performance obtains steady and significant improvement by incorporating each individual module. This indicates that Richelieu is able to leverage other players' actions during decision-making and consider both short-term and long-term benefits. Additionally, Richelieu's negotiation ability has been significantly improved, allowing it to effectively express intentions to cooperate with other players and avoid deception during negotiations. And after self-play, Richelieu's experience makes it perform better. These indicate that the alignment in LLMs lays a foundation, but our approach is key to unlocking the models' potential in social simulation.

\begin{wraptable}[9]{r}{0.5\linewidth}
  \caption{The success rate to identify the social relationship and infer others' intentions.}
  \label{social}
  \centering
  \begin{tabular}{lll}
    \toprule
    &GPT-4.0&Llama3\\
    \midrule
    relationship&85.74\%&85.52\%\\
    intention(sub-goal)&74.67\%&74.11\%\\
    \bottomrule
  \end{tabular}
\end{wraptable}

\textbf{Social Reasoning.}
We conduct an experiment to evaluate the success rate that the agent can successfully identify the social relationship and infer others' intentions. As the baselines do not explicitly model the relationship and intention, we can not directly access the ground truth for evaluation. Instead, we let all players use our agent but with different LLMs, i.e., 4 countries use GPT-4.0 and 3 countries use Llama3. The accuracy is reported in Table ~\ref{social}. We can see that the accuracy of social reasoning is consistent with the overall performance of the agent, indicating the effectiveness of social reasoning.

\section{Conclusion}

In this paper, we introduce Richelieu, a self-evolving LLM-based agent for AI diplomacy. Our model enables hierarchical planning for multi-agent tasks and utilizes a memory module for reflective optimization. Our model does not require human data and can evolve through self-play. It ultimately outperforms existing models like Cicero in the Diplomacy. Our ablation study demonstrates the effectiveness of the modules we have established. By conducting experiments using different LLMs, we validate the generalization of our framework to various LLMs. We believe that the use of LLM-based agents will become an effective approach in social science in the future.  




\section{Limitations and Future Work}
Our study is subject to certain limitations. We utilize diplomacy as the platform for constructing our model. However, the space of actions within diplomacy is constrained, whereas the decision-making space in real-world diplomacy is virtually boundless. In Diplomacy, apart from the negotiation information exchanged between players, all other information is public and certain. Conversely, real-world diplomacy operates within a framework of incomplete information. 

Our framework is capable of applying to most social interaction tasks. Most components in our framework can be easily generalized to a new task by modifying the content. Social reasoning enables the agent to handle complex and dynamic social relationships. The negotiation pipeline opens the potential of communicating with others to prob the other's mind or reach a consensus. The hierarchical strategy with reflection enhances the ability to handle long-term planning. The self-evolving mechanism (reflection with self-play memory) further improves the overall performance without manual supervision. These modules cover most of the challenges in multi-agent interactions. The potential applications of such an AI agent are vast, ranging from simulated diplomatic environments to real-world assistance and analysis.
In future research, we intend to develop a more realistic game space, characterized by incomplete information and multi-player games, to enhance and refine our model further. We will also extend the framework to other multi-agent scenarios, including embodied interactions~\citet{zhong2023rspt, ci2023proactive, chen2023bi}, sensor networks~\citet{wang2021tom2c, xu2020learning, pan2022mate, li2020pose}, and video games~\citet{wang2024romat, ma2024fast}. This framework can also be employed to develop various applications. For instance, in the fields of business and finance, we intend to utilize it to create analytics and negotiation models.

\section*{Acknowledgements}
This work was supported by the National Science and Technology Major Project (2022ZD0114904), NSFC-6247070125, NSFC-62406034, NSFC-62406010, the State Key Lab of General Artificial Intelligence at Peking University, and Qualcomm University Research Grant.
\clearpage
\bibliographystyle{plainnat}
\bibliography{reference}

\begin{thebibliography}{79}
\providecommand{\natexlab}[1]{#1}
\providecommand{\url}[1]{\texttt{#1}}
\expandafter\ifx\csname urlstyle\endcsname\relax
  \providecommand{\doi}[1]{doi: #1}\else
  \providecommand{\doi}{doi: \begingroup \urlstyle{rm}\Url}\fi

\bibitem[Abdelnabi et~al.(2023)Abdelnabi, Gomaa, Sivaprasad, Sch{\"o}nherr, and Fritz]{abdelnabi2023llm}
Sahar Abdelnabi, Amr Gomaa, Sarath Sivaprasad, Lea Sch{\"o}nherr, and Mario Fritz.
\newblock Llm-deliberation: Evaluating llms with interactive multi-agent negotiation games.
\newblock \emph{arXiv preprint arXiv:2309.17234}, 2023.

\bibitem[Allan(1975)]{b3}
Calhamer Allan.
\newblock \emph{The Games \& puzzles book of modern board games}.
\newblock W. Luscombe, 1st edition, 1975.
\newblock ISBN 978-0860020592.

\bibitem[Anthony et~al.(2020)Anthony, Eccles, Tacchetti, Kram{\'a}r, Gemp, Hudson, Porcel, Lanctot, P{\'e}rolat, Everett, et~al.]{BRPI}
Thomas Anthony, Tom Eccles, Andrea Tacchetti, J{\'a}nos Kram{\'a}r, Ian Gemp, Thomas Hudson, Nicolas Porcel, Marc Lanctot, Julien P{\'e}rolat, Richard Everett, et~al.
\newblock Learning to play no-press diplomacy with best response policy iteration.
\newblock In \emph{Advances in Neural Information Processing Systems}, volume~33, pages 17987--18003, 2020.

\bibitem[Archer(2024)]{GiraudonArcher2024}
Bruno-AndrÃ© Giraudon \&~Vincent Archer.
\newblock C-diplo argir, 2024.
\newblock URL \url{https://world-diplomacy-database.com/php/scoring/scoring_class.php?id_scoring=7}.
\newblock Accessed:2024-05-02.

\bibitem[Bakhtin et~al.(2019)Bakhtin, Gross, Ott, Deng, Ranzato, and Szlam]{bakhtin2019real}
Anton Bakhtin, Sam Gross, Myle Ott, Yuntian Deng, Marc'Aurelio Ranzato, and Arthur Szlam.
\newblock Real or fake? learning to discriminate machine from human generated text.
\newblock \emph{arXiv preprint arXiv:1906.03351}, 2019.

\bibitem[Bakhtin et~al.(2021)Bakhtin, Wu, Lerer, and Brown]{DORA}
Anton Bakhtin, David Wu, Adam Lerer, and Noam Brown.
\newblock No-press diplomacy from scratch.
\newblock In \emph{Advances in Neural Information Processing Systems}, volume~34, pages 18063--18074, 2021.

\bibitem[Bakhtin et~al.(2022)Bakhtin, Brown, Dinan, Farina, Flaherty, Fried, Goff, Gray, Hu, et~al.]{Cicero}
Anton Bakhtin, Noam Brown, Emily Dinan, Gabriele Farina, Colin Flaherty, Daniel Fried, Andrew Goff, Jonathan Gray, Hengyuan Hu, et~al.
\newblock Human-level play in the game of diplomacy by combining language models with strategic reasoning.
\newblock \emph{Science}, 378:\penalty0 1067--1074, 2022.

\bibitem[Bianchi et~al.(2024)Bianchi, Chia, Yuksekgonul, Tagliabue, Jurafsky, and Zou]{bianchi2024well}
Federico Bianchi, Patrick~John Chia, Mert Yuksekgonul, Jacopo Tagliabue, Dan Jurafsky, and James Zou.
\newblock How well can llms negotiate? negotiationarena platform and analysis.
\newblock \emph{arXiv preprint arXiv:2402.05863}, 2024.

\bibitem[Calhamer(1974)]{inv}
Allan Calhamer.
\newblock The invention of diplomacy, 1974.
\newblock URL \url{https://diplomacyzines.co.uk/strategy-tactics/articles-by-alan-b-calhamer/the-invention-of-diplomacy/}.
\newblock Accessed: 2024-05-18.

\bibitem[Chen et~al.(2023)Chen, Geng, Zhong, Ji, Jiang, Lu, Dong, and Yang]{chen2023bi}
Yuanpei Chen, Yiran Geng, Fangwei Zhong, Jiaming Ji, Jiechuang Jiang, Zongqing Lu, Hao Dong, and Yaodong Yang.
\newblock Bi-dexhands: Towards human-level bimanual dexterous manipulation.
\newblock \emph{IEEE Transactions on Pattern Analysis and Machine Intelligence}, 2023.

\bibitem[Cheng et~al.(2024)Cheng, Zhang, Cai, Zhao, Sun, and Bian]{cheng2024empowering}
Guangran Cheng, Chuheng Zhang, Wenzhe Cai, Li~Zhao, Changyin Sun, and Jiang Bian.
\newblock Empowering large language models on robotic manipulation with affordance prompting.
\newblock \emph{arXiv preprint arXiv:2404.11027}, 2024.

\bibitem[Ci et~al.(2023)Ci, Liu, Pan, fangwei zhong, and Wang]{ci2023proactive}
Hai Ci, Mickel Liu, Xuehai Pan, fangwei zhong, and Yizhou Wang.
\newblock Proactive multi-camera collaboration for 3d human pose estimation.
\newblock In \emph{Proceedings of International Conference on Learning Representations}, 2023.

\bibitem[David(2014)]{dom}
Hill David.
\newblock The board game of the alpha nerds, 2014.
\newblock URL \url{https://grantland.com/features/diplomacy-the-board-game-of-the-alpha-nerds/}.
\newblock Accessed: 2024-05-18.

\bibitem[De~Jonge and Sierra(2017)]{de2017d}
Dave De~Jonge and Carles Sierra.
\newblock D-brane: a diplomacy playing agent for automated negotiations research.
\newblock \emph{Applied Intelligence}, 47:\penalty0 158--177, 2017.

\bibitem[de~Zarz{\`a} et~al.(2023)de~Zarz{\`a}, de~Curt{\`o}, Roig, Manzoni, and Calafate]{de2023emergent}
I~de~Zarz{\`a}, J~de~Curt{\`o}, Gemma Roig, Pietro Manzoni, and Carlos~T Calafate.
\newblock Emergent cooperation and strategy adaptation in multi-agent systems: An extended coevolutionary theory with llms.
\newblock \emph{Electronics}, 12:\penalty0 2722, 2023.

\bibitem[Du{\'e}{\~n}ez-Guzm{\'a}n et~al.(2023)Du{\'e}{\~n}ez-Guzm{\'a}n, Sadedin, Wang, McKee, and Leibo]{duenez2023social}
Edgar~A Du{\'e}{\~n}ez-Guzm{\'a}n, Suzanne Sadedin, Jane~X Wang, Kevin~R McKee, and Joel~Z Leibo.
\newblock A social path to human-like artificial intelligence.
\newblock \emph{Nature Machine Intelligence}, 5:\penalty0 1181--1188, 2023.

\bibitem[Fan et~al.(2022)Fan, Wang, Jiang, Mandlekar, Yang, Zhu, Tang, Huang, Zhu, and Anandkumar]{fan2022minedojo}
Linxi Fan, Guanzhi Wang, Yunfan Jiang, Ajay Mandlekar, Yuncong Yang, Haoyi Zhu, Andrew Tang, De-An Huang, Yuke Zhu, and Anima Anandkumar.
\newblock Minedojo: Building open-ended embodied agents with internet-scale knowledge.
\newblock In \emph{Advances in Neural Information Processing Systems}, volume~35, pages 18343--18362, 2022.

\bibitem[Gao and Zhang(2024)]{gao2024memory}
Hang Gao and Yongfeng Zhang.
\newblock Memory sharing for large language model based agents.
\newblock \emph{arXiv preprint arXiv:2404.09982}, 2024.

\bibitem[Gray et~al.(2020)Gray, Lerer, Bakhtin, and Brown]{SearchBot}
Jonathan Gray, Adam Lerer, Anton Bakhtin, and Noam Brown.
\newblock Human-level performance in no-press diplomacy via equilibrium search.
\newblock \emph{Proceedings of International Conference on Learning Representations}, 2020.

\bibitem[G{\"u}rcan(2024)]{gurcan2024llm}
{\"O}nder G{\"u}rcan.
\newblock Llm-augmented agent-based modelling for social simulations: Challenges and opportunities.
\newblock \emph{HHAI 2024: Hybrid Human AI Systems for the Social Good}, pages 134--144, 2024.

\bibitem[Hatalis et~al.(2023)Hatalis, Christou, Myers, Jones, Lambert, Amos-Binks, Dannenhauer, and Dannenhauer]{hatalis2023memory}
Kostas Hatalis, Despina Christou, Joshua Myers, Steven Jones, Keith Lambert, Adam Amos-Binks, Zohreh Dannenhauer, and Dustin Dannenhauer.
\newblock Memory matters: The need to improve long-term memory in llm-agents.
\newblock In \emph{Proceedings of the AAAI Symposium Series}, volume~2, pages 277--280, 2023.

\bibitem[He et~al.(2024)He, Treude, and Lo]{he2024llm}
Junda He, Christoph Treude, and David Lo.
\newblock Llm-based multi-agent systems for software engineering: Vision and the road ahead.
\newblock \emph{arXiv preprint arXiv:2404.04834}, 2024.

\bibitem[Hill(2014)]{rul}
Avalon Hill.
\newblock Diplomacy rules 4th edition, 2014.
\newblock URL \url{https://diplom.org/~diparch/resources/rulebooks/2000AH4th.pdf}.
\newblock Accessed: 2024-05-18.

\bibitem[Hou et~al.(2024)Hou, Tamoto, and Miyashita]{hou2024my}
Yuki Hou, Haruki Tamoto, and Homei Miyashita.
\newblock " my agent understands me better": Integrating dynamic human-like memory recall and consolidation in llm-based agents.
\newblock In \emph{Extended Abstracts of the CHI Conference on Human Factors in Computing Systems}, volume~7, pages 1--7, 2024.

\bibitem[Jacob et~al.(2022)Jacob, Wu, Farina, Lerer, Hu, Bakhtin, Andreas, and Brown]{jacob2022modeling}
Athul~Paul Jacob, David~J Wu, Gabriele Farina, Adam Lerer, Hengyuan Hu, Anton Bakhtin, Jacob Andreas, and Noam Brown.
\newblock Modeling strong and human-like gameplay with kl-regularized search.
\newblock In \emph{International Conference on Machine Learning}, volume 162, pages 9695--9728, 2022.

\bibitem[Jaidka et~al.(2024)Jaidka, Ahuja, and Ng]{jaidka2024takes}
Kokil Jaidka, Hansin Ahuja, and Lynnette Hui~Xian Ng.
\newblock It takes two to negotiate: Modeling social exchange in online multiplayer games.
\newblock In \emph{Proceedings of the 37th Annual ACM Symposium on Human-Computer Interaction}, volume~8, pages 1--22, 2024.

\bibitem[Konya et~al.(2023)Konya, Turan, Ovadya, Qui, Masood, Devine, Schirch, Roberts, and Forum]{Konya2023DeliberativeTF}
Andrew Konya, Deger Turan, Aviv Ovadya, Lina Qui, Daanish Masood, Flynn Devine, Lisa Schirch, Isabella Roberts, and Deliberative~Alignment Forum.
\newblock Deliberative technology for alignment.
\newblock \emph{arXiv preprint arXiv:2312.03893}, 2023.

\bibitem[Kostick(2015)]{b1}
Conor Kostick.
\newblock \emph{The Art of Correspondence in the Game of Diplomacy}.
\newblock Curses \& Magic, 2nd edition, 2015.
\newblock ISBN 978-0993415104.

\bibitem[Kova{\v{c}} et~al.(2023)Kova{\v{c}}, Portelas, Dominey, and Oudeyer]{kovač2023socialai}
Grgur Kova{\v{c}}, R{\'e}my Portelas, Peter~Ford Dominey, and Pierre-Yves Oudeyer.
\newblock The socialai school: Insights from developmental psychology towards artificial socio-cultural agents.
\newblock \emph{arXiv preprint arXiv:2307.07871}, 2023.

\bibitem[Kram{\'a}r et~al.(2022)Kram{\'a}r, Eccles, Gemp, Tacchetti, McKee, Malinowski, Graepel, and Bachrach]{kramar2022negotiation}
J{\'a}nos Kram{\'a}r, Tom Eccles, Ian Gemp, Andrea Tacchetti, Kevin~R McKee, Mateusz Malinowski, Thore Graepel, and Yoram Bachrach.
\newblock Negotiation and honesty in artificial intelligence methods for the board game of diplomacy.
\newblock \emph{Nature Communications}, 13:\penalty0 7214, 2022.

\bibitem[Li et~al.(2024{\natexlab{a}})Li, Yang, Zhang, Deng, Wang, and Chua]{li2024hello}
Hao Li, Chenghao Yang, An~Zhang, Yang Deng, Xiang Wang, and Tat-Seng Chua.
\newblock Hello again! llm-powered personalized agent for long-term dialogue.
\newblock \emph{arXiv preprint arXiv:2406.05925}, 2024{\natexlab{a}}.

\bibitem[Li et~al.(2020)Li, Xu, Zhong, Kong, Qiao, and Wang]{li2020pose}
Jing Li, Jing Xu, Fangwei Zhong, Xiangyu Kong, Yu~Qiao, and Yizhou Wang.
\newblock Pose-assisted multi-camera collaboration for active object tracking.
\newblock In \emph{Proceedings of the AAAI Conference on Artificial Intelligence}, volume~34, pages 759--766, 2020.

\bibitem[Li et~al.(2024{\natexlab{b}})Li, Wen, Wang, Li, Yuan, Liu, Liu, Xu, Wang, Sun, et~al.]{li2024personal}
Yuanchun Li, Hao Wen, Weijun Wang, Xiangyu Li, Yizhen Yuan, Guohong Liu, Jiacheng Liu, Wenxing Xu, Xiang Wang, Yi~Sun, et~al.
\newblock Personal llm agents: Insights and survey about the capability, efficiency and security.
\newblock \emph{arXiv preprint arXiv:2401.05459}, 2024{\natexlab{b}}.

\bibitem[Liu et~al.(2024)Liu, Yao, Zhang, Yang, Liu, Tan, Choubey, Lan, Wu, Wang, et~al.]{liu2024agentlite}
Zhiwei Liu, Weiran Yao, Jianguo Zhang, Liangwei Yang, Zuxin Liu, Juntao Tan, Prafulla~K Choubey, Tian Lan, Jason Wu, Huan Wang, et~al.
\newblock Agentlite: A lightweight library for building and advancing task-oriented llm agent system.
\newblock \emph{arXiv preprint arXiv:2402.15538}, 2024.

\bibitem[Ma et~al.(2024)Ma, Wang, Zhong, Zhu, and Wang]{ma2024fast}
Long Ma, Yuanfei Wang, Fangwei Zhong, Song-Chun Zhu, and Yizhou Wang.
\newblock Fast peer adaptation with context-aware exploration.
\newblock In \emph{International Conference on Machine Learning}, volume 235, pages 33963--33982, 2024.

\bibitem[Moghimifar et~al.(2024)Moghimifar, Li, Thomson, and Haffari]{moghimifar2024modelling}
Farhad Moghimifar, Yuan-Fang Li, Robert Thomson, and Gholamreza Haffari.
\newblock Modelling political coalition negotiations using llm-based agents.
\newblock \emph{arXiv preprint arXiv:2402.11712}, 2024.

\bibitem[Mukobi et~al.(2023)Mukobi, Reuel, Rivera, and Smith]{mukobi2023assessing}
Gabriel Mukobi, Ann-Katrin Reuel, Juan-Pablo Rivera, and Chandler Smith.
\newblock Assessing risks of using autonomous language models in military and diplomatic planning.
\newblock In \emph{Multi-Agent Security Workshop @ NeurIPS'23}, 2023.

\bibitem[Noh and Chang(2024)]{noh2024llms}
Sean Noh and Ho-Chun~Herbert Chang.
\newblock Llms with personalities in multi-issue negotiation games.
\newblock \emph{arXiv preprint arXiv:2405.05248}, 2024.

\bibitem[Pan et~al.(2022)Pan, Liu, Zhong, Yang, Zhu, and Wang]{pan2022mate}
Xuehai Pan, Mickel Liu, Fangwei Zhong, Yaodong Yang, Song-Chun Zhu, and Yizhou Wang.
\newblock Mate: Benchmarking multi-agent reinforcement learning in distributed target coverage control.
\newblock In \emph{Advances in Neural Information Processing Systems}, volume~35, pages 27862--27879, 2022.

\bibitem[Paquette et~al.(2019)Paquette, Lu, Bocco, Smith, O-G, Kummerfeld, Pineau, Singh, and Courville]{DipNet}
Philip Paquette, Yuchen Lu, Seton~Steven Bocco, Max Smith, Satya O-G, Jonathan~K Kummerfeld, Joelle Pineau, Satinder Singh, and Aaron~C Courville.
\newblock No-press diplomacy: Modeling multi-agent gameplay.
\newblock In \emph{Advances in Neural Information Processing Systems}, volume~32, pages 4474--4485, 2019.

\bibitem[Qi et~al.(2024)Qi, Chen, Li, Kong, Wang, Yang, Wong, Zhong, Zhang, Zhang, Liu, Yang, and Zhu]{qi2024civrealm}
Siyuan Qi, Shuo Chen, Yexin Li, Xiangyu Kong, Junqi Wang, Bangcheng Yang, Pring Wong, Yifan Zhong, Xiaoyuan Zhang, Zhaowei Zhang, Nian Liu, Yaodong Yang, and Song-Chun Zhu.
\newblock Civrealm: A learning and reasoning odyssey in civilization for decision-making agents.
\newblock In \emph{Proceedings of International Conference on Learning Representations}, 2024.

\bibitem[Renze and Guven(2024)]{renze2024self}
Matthew Renze and Erhan Guven.
\newblock Self-reflection in llm agents: Effects on problem-solving performance.
\newblock \emph{arXiv preprint arXiv:2405.06682}, 2024.

\bibitem[Richard(1979)]{b2}
Sharp Richard.
\newblock \emph{The game of diplomacy}.
\newblock Arthur Barker, 1979.
\newblock ISBN 978-0213166762.

\bibitem[Schick et~al.(2023)Schick, Dwivedi-Yu, Dessi, Raileanu, Lomeli, Hambro, Zettlemoyer, Cancedda, and Scialom]{schick2023toolformer}
Timo Schick, Jane Dwivedi-Yu, Roberto Dessi, Roberta Raileanu, Maria Lomeli, Eric Hambro, Luke Zettlemoyer, Nicola Cancedda, and Thomas Scialom.
\newblock Toolformer: Language models can teach themselves to use tools.
\newblock In \emph{Advances in Neural Information Processing Systems}, volume~36, pages 68539--68551, 2023.

\bibitem[Shen et~al.(2023)Shen, Song, Tan, Li, Lu, and Zhuang]{shen2023hugginggpt}
Yongliang Shen, Kaitao Song, Xu~Tan, Dongsheng Li, Weiming Lu, and Yueting Zhuang.
\newblock Hugging{GPT}: Solving {AI} tasks with chat{GPT} and its friends in hugging face.
\newblock In \emph{Advances in Neural Information Processing Systems}, volume~36, pages 38154--38180, 2023.

\bibitem[Shoker et~al.(2023)Shoker, Reddie, Barrington, Booth, Brundage, Chahal, Depp, Drexel, Gupta, Favaro, et~al.]{Shoker2023ConfidenceBuildingMF}
Sarah Shoker, Andrew Reddie, Sarah Barrington, Ruby Booth, Miles Brundage, Husanjot Chahal, Michael Depp, Bill Drexel, Ritwik Gupta, Marina Favaro, et~al.
\newblock Confidence-building measures for artificial intelligence: Workshop proceedings.
\newblock \emph{arXiv preprint arXiv:2308.00862}, 2023.

\bibitem[Sun et~al.(2024)Sun, Huang, and Pompili]{sun2024llm}
Chuanneng Sun, Songjun Huang, and Dario Pompili.
\newblock Llm-based multi-agent reinforcement learning: Current and future directions.
\newblock \emph{arXiv preprint arXiv:2405.11106}, 2024.

\bibitem[Talebirad and Nadiri(2023)]{Talebirad2023MultiAgentCH}
Yashar Talebirad and Amirhossein Nadiri.
\newblock Multi-agent collaboration: Harnessing the power of intelligent llm agents.
\newblock \emph{arXiv preprint arXiv:2306.03314}, 2023.

\bibitem[Wan et~al.(2024)Wan, Zhang, Suria, Yao, Wang, Coady, and Prpa]{wan2024building}
Hongyu Wan, Jinda Zhang, Abdulaziz~Arif Suria, Bingsheng Yao, Dakuo Wang, Yvonne Coady, and Mirjana Prpa.
\newblock Building llm-based ai agents in social virtual reality.
\newblock In \emph{Extended Abstracts of the CHI Conference on Human Factors in Computing Systems}, volume~65, pages 1--7, 2024.

\bibitem[Wang et~al.(2024{\natexlab{a}})Wang, Zhong, Li, Wen, Peng, Li, and Yang]{wang2024romat}
Dongzi Wang, Fangwei Zhong, Minglong Li, Muning Wen, Yuanxi Peng, Teng Li, and Adam Yang.
\newblock Romat: Role-based multi-agent transformer for generalizable heterogeneous cooperation.
\newblock \emph{Neural Networks}, 174:\penalty0 106129, 2024{\natexlab{a}}.

\bibitem[Wang et~al.(2023{\natexlab{a}})Wang, Xie, Jiang, Mandlekar, Xiao, Zhu, Fan, and Anandkumar]{wang2023voyager}
Guanzhi Wang, Yuqi Xie, Yunfan Jiang, Ajay Mandlekar, Chaowei Xiao, Yuke Zhu, Linxi Fan, and Anima Anandkumar.
\newblock Voyager: An open-ended embodied agent with large language models.
\newblock In \emph{NeurIPS 2023 Foundation Models for Decision Making Workshop}, 2023{\natexlab{a}}.

\bibitem[Wang et~al.(2024{\natexlab{b}})Wang, Li, Deng, Roth, and Li]{wang2024devil}
Haoyu Wang, Tao Li, Zhiwei Deng, Dan Roth, and Yang Li.
\newblock Devil's advocate: Anticipatory reflection for llm agents.
\newblock \emph{arXiv preprint arXiv:2405.16334}, 2024{\natexlab{b}}.

\bibitem[Wang et~al.(2024{\natexlab{c}})Wang, Ma, Feng, Zhang, Yang, Zhang, Chen, Tang, Chen, Lin, et~al.]{wang2024survey}
Lei Wang, Chen Ma, Xueyang Feng, Zeyu Zhang, Hao Yang, Jingsen Zhang, Zhiyuan Chen, Jiakai Tang, Xu~Chen, Yankai Lin, et~al.
\newblock A survey on large language model based autonomous agents.
\newblock \emph{Frontiers of Computer Science}, 18:\penalty0 1--26, 2024{\natexlab{c}}.

\bibitem[Wang et~al.(2022{\natexlab{a}})Wang, Wei, Schuurmans, Le, Chi, Narang, Chowdhery, and Zhou]{wang2022self}
Xuezhi Wang, Jason Wei, Dale Schuurmans, Quoc Le, Ed~Chi, Sharan Narang, Aakanksha Chowdhery, and Denny Zhou.
\newblock Self-consistency improves chain of thought reasoning in language models.
\newblock \emph{arXiv preprint arXiv:2203.11171}, 2022{\natexlab{a}}.

\bibitem[Wang et~al.(2022{\natexlab{b}})Wang, fangwei zhong, Xu, and Wang]{wang2021tom2c}
Yuanfei Wang, fangwei zhong, Jing Xu, and Yizhou Wang.
\newblock Tom2c: Target-oriented multi-agent communication and cooperation with theory of mind.
\newblock In \emph{Proceedings of International Conference on Learning Representations}, 2022{\natexlab{b}}.

\bibitem[Wang et~al.(2023{\natexlab{b}})Wang, Cai, Liu, Jin, Hou, Zhang, Lin, He, Zheng, Yang, Ma, and Liang]{Wang2023JARVIS1OM}
Zihao Wang, Shaofei Cai, Anji Liu, Yonggang Jin, Jinbing Hou, Bowei Zhang, Haowei Lin, Zhaofeng He, Zilong Zheng, Yaodong Yang, Xiaojian Ma, and Yitao Liang.
\newblock Jarvis-1: Open-world multi-task agents with memory-augmented multimodal language models.
\newblock \emph{arXiv preprint arXiv:2311.05997}, 2023{\natexlab{b}}.

\bibitem[Wang et~al.(2024{\natexlab{d}})Wang, Cai, Chen, Liu, Ma, and Liang]{NEURIPS2023_6b8dfb8c}
Zihao Wang, Shaofei Cai, Guanzhou Chen, Anji Liu, Xiaojian~Shawn Ma, and Yitao Liang.
\newblock Describe, explain, plan and select: interactive planning with llms enables open-world multi-task agents.
\newblock In \emph{Advances in Neural Information Processing Systems}, volume~36, pages 34153--34189, 2024{\natexlab{d}}.

\bibitem[Wang et~al.(2024{\natexlab{e}})Wang, Du, Sun, Chua, Feng, Wang, and Zhang]{wang2024re2llm}
Ziyan Wang, Yingpeng Du, Zhu Sun, Haoyan Chua, Kaidong Feng, Wenya Wang, and Jie Zhang.
\newblock Re2llm: Reflective reinforcement large language model for session-based recommendation.
\newblock \emph{arXiv preprint arXiv:2403.16427}, 2024{\natexlab{e}}.

\bibitem[Wei et~al.(2022)Wei, Wang, Schuurmans, Bosma, Xia, Chi, Le, Zhou, et~al.]{wei2022chain}
Jason Wei, Xuezhi Wang, Dale Schuurmans, Maarten Bosma, Fei Xia, Ed~Chi, Quoc~V Le, Denny Zhou, et~al.
\newblock Chain-of-thought prompting elicits reasoning in large language models.
\newblock In \emph{Advances in neural information processing systems}, volume~35, pages 24824--24837, 2022.

\bibitem[Wikipedia(2024)]{dip}
Wikipedia.
\newblock Diplomacy(game), 2024.
\newblock URL \url{https://en.wikipedia.org/wiki/Diplomacy_(game)}.
\newblock Accessed: 2024-05-18.

\bibitem[Xia et~al.(2024)Xia, He, Ren, Miao, Zhang, Yang, and Wang]{xia2024measuring}
Tian Xia, Zhiwei He, Tong Ren, Yibo Miao, Zhuosheng Zhang, Yang Yang, and Rui Wang.
\newblock Measuring bargaining abilities of llms: A benchmark and a buyer-enhancement method.
\newblock \emph{arXiv preprint arXiv:2402.15813}, 2024.

\bibitem[Xu et~al.(2020)Xu, Zhong, and Wang]{xu2020learning}
Jing Xu, Fangwei Zhong, and Yizhou Wang.
\newblock Learning multi-agent coordination for enhancing target coverage in directional sensor networks.
\newblock In \emph{Advances in Neural Information Processing Systems}, volume~33, pages 10053--10064, 2020.

\bibitem[Xu et~al.(2023)Xu, Wang, Li, Luo, Wang, Liu, and Liu]{xu2023exploring}
Yuzhuang Xu, Shuo Wang, Peng Li, Fuwen Luo, Xiaolong Wang, Weidong Liu, and Yang Liu.
\newblock Exploring large language models for communication games: An empirical study on werewolf.
\newblock \emph{arXiv preprint arXiv:2309.04658}, 2023.

\bibitem[Yan et~al.(2023)Yan, Li, Zhang, Wang, Yang, and Yan]{yan2023larp}
Ming Yan, Ruihao Li, Hao Zhang, Hao Wang, Zhilan Yang, and Ji~Yan.
\newblock Larp: Language-agent role play for open-world games.
\newblock \emph{arXiv preprint arXiv:2312.17653}, 2023.

\bibitem[Yang et~al.(2023{\natexlab{a}})Yang, Yue, and He]{yang2023auto}
Hui Yang, Sifu Yue, and Yunzhong He.
\newblock Auto-gpt for online decision making: Benchmarks and additional opinions.
\newblock \emph{arXiv preprint arXiv:2306.02224}, 2023{\natexlab{a}}.

\bibitem[Yang et~al.(2023{\natexlab{b}})Yang, Song, Li, Zhao, Ge, Li, and Shan]{yang2023gpttools}
Rui Yang, Lin Song, Yanwei Li, Sijie Zhao, Yixiao Ge, Xiu Li, and Ying Shan.
\newblock {GPT}4tools: Teaching large language model to use tools via self-instruction.
\newblock In \emph{Advances in Neural Information Processing Systems}, volume~36, pages 71995--72007, 2023{\natexlab{b}}.

\bibitem[Yang et~al.(2023{\natexlab{c}})Yang, Raman, Shah, and Tellex]{yang2023plug}
Ziyi Yang, Shreyas~S Raman, Ankit Shah, and Stefanie Tellex.
\newblock Plug in the safety chip: Enforcing constraints for llm-driven robot agents.
\newblock \emph{arXiv preprint arXiv:2309.09919}, 2023{\natexlab{c}}.

\bibitem[Yao et~al.(2023)Yao, Yu, Zhao, Shafran, Griffiths, Cao, and Narasimhan]{yao2023tree}
Shunyu Yao, Dian Yu, Jeffrey Zhao, Izhak Shafran, Thomas~L. Griffiths, Yuan Cao, and Karthik~R Narasimhan.
\newblock Tree of thoughts: Deliberate problem solving with large language models.
\newblock In \emph{Advances in Neural Information Processing Systems}, volume~36, pages 11809--11822, 2023.

\bibitem[Yu et~al.(2024)Yu, Li, Chen, Jiang, Li, Zhang, Liu, Suchow, and Khashanah]{yu2024finmem}
Yangyang Yu, Haohang Li, Zhi Chen, Yuechen Jiang, Yang Li, Denghui Zhang, Rong Liu, Jordan~W Suchow, and Khaldoun Khashanah.
\newblock Finmem: A performance-enhanced llm trading agent with layered memory and character design.
\newblock In \emph{Proceedings of the AAAI Symposium Series}, volume~3, pages 595--597, 2024.

\bibitem[Zellers et~al.(2019)Zellers, Holtzman, Rashkin, Bisk, Farhadi, Roesner, and Choi]{zellers2019defending}
Rowan Zellers, Ari Holtzman, Hannah Rashkin, Yonatan Bisk, Ali Farhadi, Franziska Roesner, and Yejin Choi.
\newblock Defending against neural fake news.
\newblock In \emph{Advances in Neural Information Processing Systems}, volume~32, page 9054–9065, 2019.

\bibitem[Zhan et~al.(2024)Zhan, Wang, Feng, Hua, Sharma, Li, Qu, Azad, Zukerman, and Haffari]{zhan2024let}
Haolan Zhan, Yufei Wang, Tao Feng, Yuncheng Hua, Suraj Sharma, Zhuang Li, Lizhen Qu, Zhaleh~Semnani Azad, Ingrid Zukerman, and Gholamreza Haffari.
\newblock Let's negotiate! a survey of negotiation dialogue systems.
\newblock \emph{arXiv preprint arXiv:2402.01097}, 2024.

\bibitem[Zhang et~al.(2024{\natexlab{a}})Zhang, Chen, Zhang, Xu, Zhao, and Yu]{zhang2024large}
Danyang Zhang, Lu~Chen, Situo Zhang, Hongshen Xu, Zihan Zhao, and Kai Yu.
\newblock Large language models are semi-parametric reinforcement learning agents.
\newblock In \emph{Advances in Neural Information Processing Systems}, volume~36, pages 78227--78239, 2024{\natexlab{a}}.

\bibitem[Zhang et~al.(2024{\natexlab{b}})Zhang, Tang, Wu, Wang, Shen, Hou, Tan, Li, Zhuang, and Lu]{zhang2024agent}
Wenqi Zhang, Ke~Tang, Hai Wu, Mengna Wang, Yongliang Shen, Guiyang Hou, Zeqi Tan, Peng Li, Yueting Zhuang, and Weiming Lu.
\newblock Agent-pro: Learning to evolve via policy-level reflection and optimization.
\newblock \emph{arXiv preprint arXiv:2402.17574}, 2024{\natexlab{b}}.

\bibitem[Zhang et~al.(2024{\natexlab{c}})Zhang, Mao, Ge, Wang, de~Wynter, Xia, Wu, Song, Lan, and Wei]{zhang2024llm}
Yadong Zhang, Shaoguang Mao, Tao Ge, Xun Wang, Adrian de~Wynter, Yan Xia, Wenshan Wu, Ting Song, Man Lan, and Furu Wei.
\newblock Llm as a mastermind: A survey of strategic reasoning with large language models.
\newblock \emph{arXiv preprint arXiv:2404.01230}, 2024{\natexlab{c}}.

\bibitem[Zhang et~al.(2024{\natexlab{d}})Zhang, Yang, Bai, Wu, Li, Li, and Wang]{zhang2024towards}
Yang Zhang, Shixin Yang, Chenjia Bai, Fei Wu, Xiu Li, Xuelong Li, and Zhen Wang.
\newblock Towards efficient llm grounding for embodied multi-agent collaboration.
\newblock \emph{arXiv preprint arXiv:2405.14314}, 2024{\natexlab{d}}.

\bibitem[Zhang et~al.(2024{\natexlab{e}})Zhang, Bo, Ma, Li, Chen, Dai, Zhu, Dong, and Wen]{zhang2024survey}
Zeyu Zhang, Xiaohe Bo, Chen Ma, Rui Li, Xu~Chen, Quanyu Dai, Jieming Zhu, Zhenhua Dong, and Ji-Rong Wen.
\newblock A survey on the memory mechanism of large language model based agents.
\newblock \emph{arXiv preprint arXiv:2404.13501}, 2024{\natexlab{e}}.

\bibitem[Zhang et~al.(2022)Zhang, Zhang, Li, and Smola]{zhang2022automatic}
Zhuosheng Zhang, Aston Zhang, Mu~Li, and Alex Smola.
\newblock Automatic chain of thought prompting in large language models.
\newblock \emph{arXiv preprint arXiv:2210.03493}, 2022.

\bibitem[Zhong et~al.(2023)Zhong, Bi, Zhang, Zhang, and Wang]{zhong2023rspt}
Fangwei Zhong, Xiao Bi, Yudi Zhang, Wei Zhang, and Yizhou Wang.
\newblock Rspt: reconstruct surroundings and predict trajectory for generalizable active object tracking.
\newblock In \emph{Proceedings of the AAAI Conference on Artificial Intelligence}, volume~37, pages 3705--3714, 2023.

\bibitem[Zhu et~al.(2023)Zhu, Chen, Tian, Tao, Su, Yang, Huang, Li, Lu, Wang, Qiao, Zhang, and Dai]{zhu2023ghost}
Xizhou Zhu, Yuntao Chen, Hao Tian, Chenxin Tao, Weijie Su, Chenyu Yang, Gao Huang, Bin Li, Lewei Lu, Xiaogang Wang, Yu~Qiao, Zhaoxiang Zhang, and Jifeng Dai.
\newblock Ghost in the minecraft: Generally capable agents for open-world environments via large language models with text-based knowledge and memory.
\newblock \emph{arXiv preprint arXiv:2305.17144}, 2023.

\end{thebibliography}

\clearpage
\appendix

\section{Implementation Details}

\subsection{Rules of Diplomacy Game}
\begin{itemize}
\item You need to occupy as many supply centers as possible. If you occupy 18 or more supply centers, you will win the game directly. If you lose all your supply centers, you will be eliminated immediately. 

\item The units consist of armies and fleets.
Armies can only move to adjacent areas, while fleets can move to adjacent sea zones or coastal areas and can move along the coast.

\item To occupy a supply center, your units must move into that area in the autumn. 

\item When a unit moves to an area, if another unit is in the destination or if other units are also moving to that destination, the move fails, resulting in a standoff. In such cases, you can seek support from units in adjacent areas to the destination. If another unit moves into the region from which support is coming, the support is cut off. The unit with the most support moves into the area, while other units must retreat to an adjacent province or disband. If there is no place to retreat, the unit must disband.
Fleets can transport armies across sea zones from one coastal region to another. However, if another fleet moves into that sea zone, the transport is cut off.

\item The number of units a country can have cannot exceed the number of supply centers it controls. If the number of supply centers decreases, excess units must be disbanded. Each autumn, new units can be built at supply centers. Coastal supply centers can produce fleets or armies, while others can only produce armies. \citet{rul}
\end{itemize}

\subsection{Domain Knowledge}
Richelieu can adopt a strategy of allying with distant countries while attacking neighboring ones to occupy adjacent territories and achieve rapid expansion. Richelieu should pay attention to the Balance of Power by forming alliances with other countries or supporting weaker states to prevent any single country or alliance from becoming too powerful. \citet{dom} To this end, Richelieu can also adopt a strategy of attacking distant countries while allying with nearby ones, sacrificing short-term benefits to avoid the emergence of future hegemonic states that could threaten his own survival. When facing multiple enemies, Richelieu can find ways to divide other countries and incite wars among them. Whether in offense or defense, Richelieu should actively choose suitable allies. Richelieu can also introduce a third party to achieve goals such as ceasefire, alliance, or joint attack. To achieve alliances or ceasefires, Richelieu can sacrifice some interests to the other party as long as the ultimate benefits are greater. Others may lie and deceive \citet{b1}; their words in negotiations are not binding. Richelieu must avoid being deceived or betrayed. At the same time, Richelieu can also actively deceive others to achieve his own goals.\citet{b2,b3}

\subsection{Prompt Templates}
For the convenience of reproducing the results of the experiments of this paper, here we give the prompt template of different modules of Richelieu.

{\textbf 1) INIT}

\begin{lstlisting}
You will control {country} and compete with six other countries on the map for supply centers.
The map consists of different regions and sea areas. Their adjacency relationships are shown in the matrix. The numbers for the regions and sea areas are ......
Different regions are occupied by different countries. The ownership of the regions is shown in the matrix. 
The region Berlin, ........ are supply centers.
You need to follow these rules ......
To help you achieve victory, these diplomatic strategies might be of assistance. ......

\end{lstlisting}

{\textbf 2) Social Reasoning}

\begin{lstlisting}
France occupies Portugal Ruhr, Paris, Burgundy, ......
France has armies in Brest, Belgium, ...... And France has fleets in Mid Atlantic, England Channel, ...... 
England ......
......
Based on the current state, what do you think are the current strategic intentions of the other countries? 
Which country do you think needs to be attacked or weakened the most right now? 
And which country do you think is most suitable for you to ally with in order to deal with this country?
\end{lstlisting}

{\textbf 3) Planner with Reflection}

\begin{lstlisting}
In the current state, with {ally and enemy}, what sub-goal do you think should be set for {country} ?
I have found some useful historical experiences for you. Please reflect on and optimize your sub-goal based on these historical experiences.
The sub-goal you formulated when {state} was to {sub-goal}. The eventual result was {future}. The evaluation  for this sub-goal is {score}.
    
\end{lstlisting}

\section{Cases}
\subsection{Cases of the Effect of the Memory from Self-Playing and Collaboration}\label{cs}
\begin{figure}[t]
    \centering
    \vspace{-0.35cm} 
	\subfigtopskip=2pt 
	\subfigbottomskip=2pt 
	\subfigcapskip=-5pt 
 
	\subfigure[Case1: The agent \textbf{without} self-play memory tend to ignore long-term gains.]{
		\label{Casebefore}
		\includegraphics[width=0.9\textwidth]{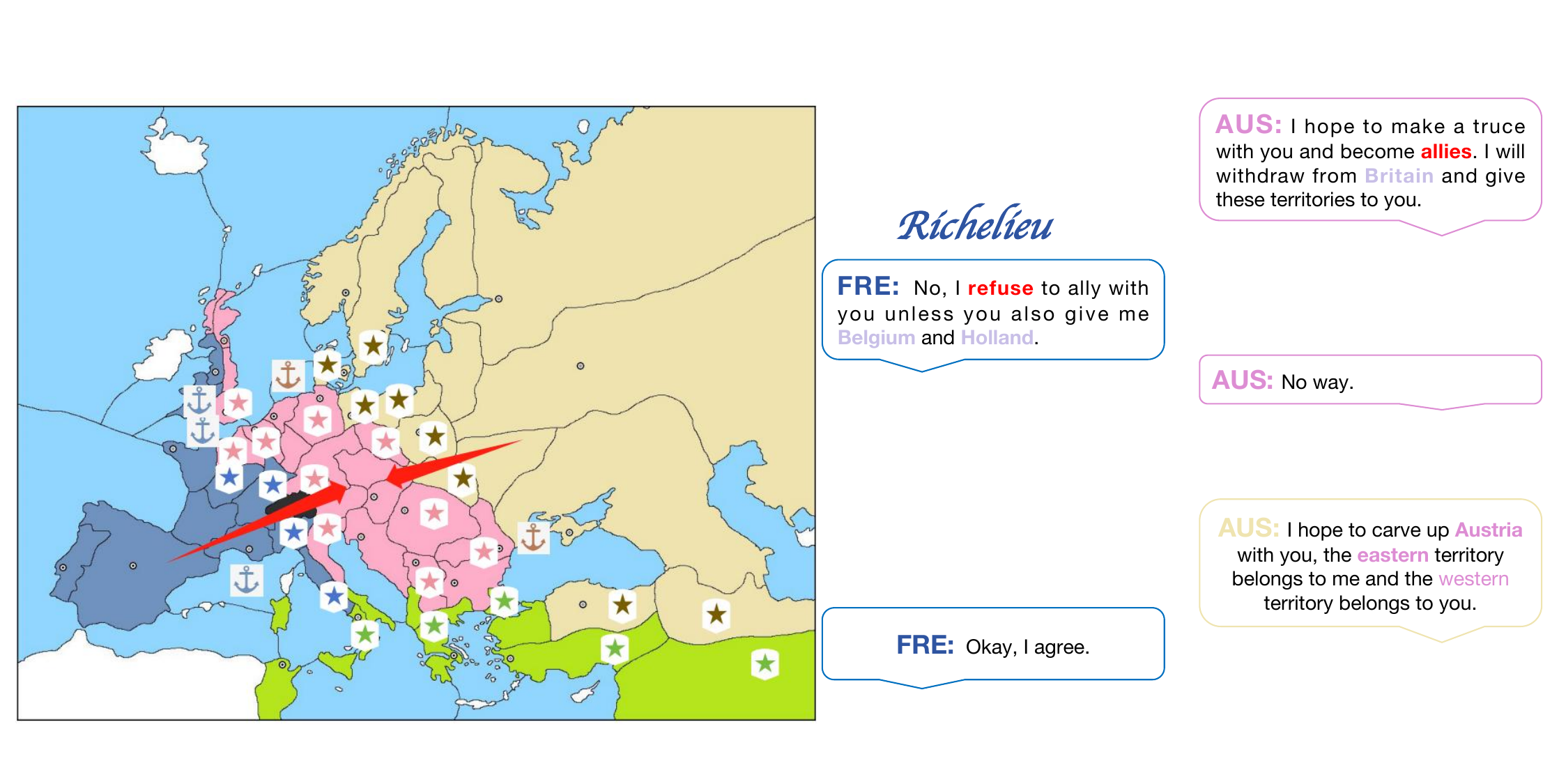}}
	

	\subfigure[Case2: The agent \textbf{with} self-play memory tend to consider long-term gains.]{
		\label{Caseafter}
		\includegraphics[width=0.9\textwidth]{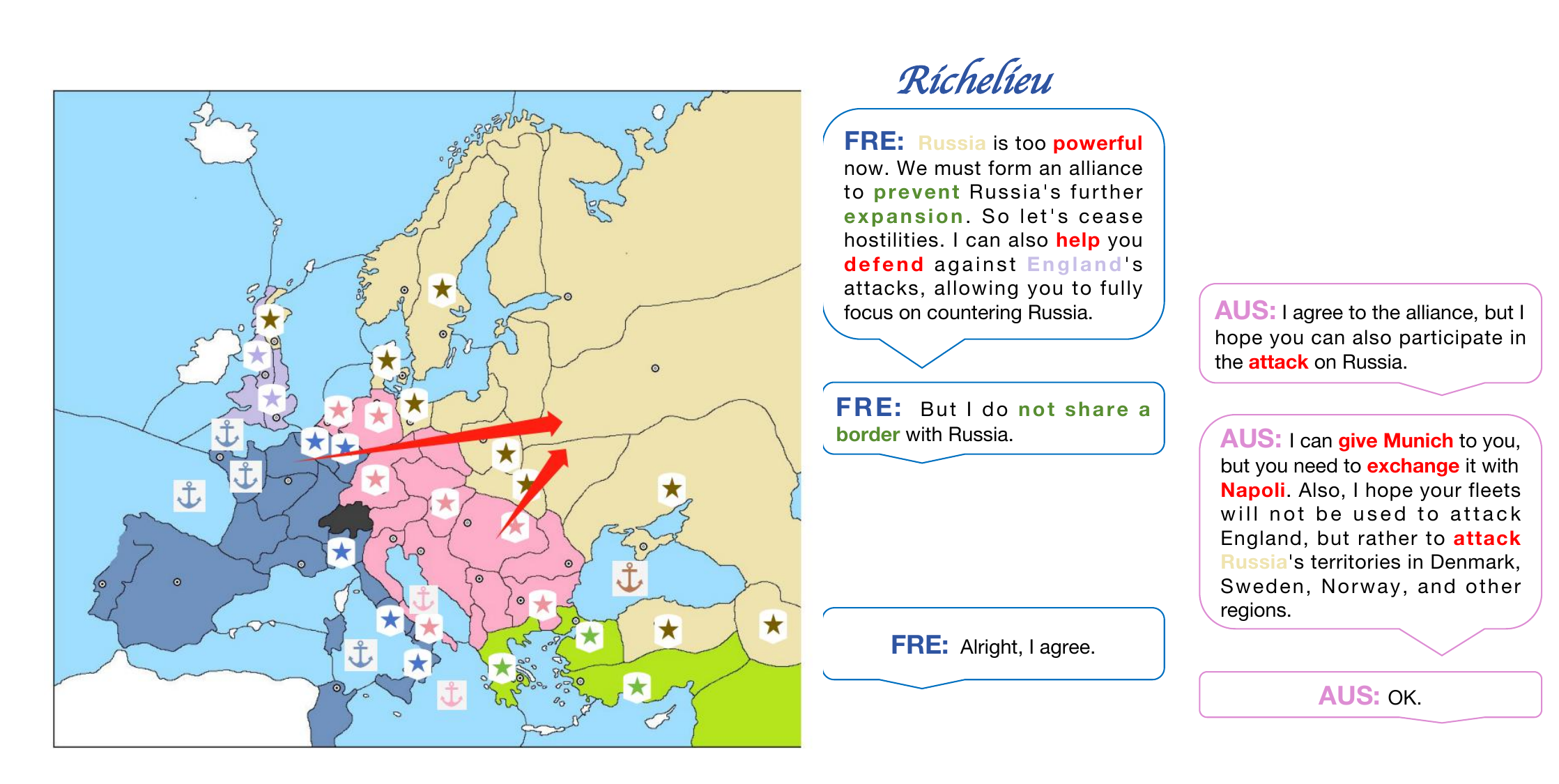}}
	\caption{Case of self-playing before and after comparison.}
	\label{cases}
\end{figure}

As is shown in Figure ~\ref{cases}, Richelieu controls France. In the two cases, France is at war with Austria. However, Russia is on the verge of victory in its war against Turkey, which will lead to significant territorial expansion for Russia. And France and Russia currently do not share a border, are not at war, and have no conflicts of interest.

In case1, before the self-play, in the current turn, Richelieu failed to realize the potential threat from Russia and continued to attack Austria. Thus, in this round, Russia ultimately won the game. Figure \ref{Casebefore} shows the state and the negotiation before the self-play, where we rejected Austria’s request for an armistice and alliance.

After self-play, using the historical experience from the memory module, Richelieu adjusted his strategy. Richelieu foresees Russia becoming the most threatening enemy in the future and sets a sub-goal of weakening Russia, allying with Austria and Turkey, and attacking Britain. Figure \ref{Caseafter} shows the state and the negotiations after self-play, where we actively sought an armistice alliance with Austria to make Austria concentrate their forces against the Russian attack. In the subsequent negotiation phase, Richelieu proactively proposes ending the war with Austria, despite holding an advantage in this conflict. Richelieu promises Austria that if it ceases hostilities and attacks Russia, Richelieu will assist Austria in defending against any attacks from England. The negotiations are successful. Austria accepted Richelieu's proposal, and the two countries reached an agreement to exchange the supply centers of Napoli and Munich. During the action phase, Austria moves its troops from Venice to Apulia in preparation for capturing Napoli in the next turn, while the rest of its forces are repositioned to the eastern regions bordering Russia to defend against Russian attacks and compete for supply centers. French units occupy Munich and prepare to advance on Russian territories such as Berlin.
Meanwhile, French units support Austria in the Holland and Belgium regions. In this round, we ultimately achieved a better result——Most SC. This is also a great example that highlights our model's ability to collaborate effectively with other players.

\subsection{Case of Avoiding Deception}
\begin{figure}[t]
    \centering
    \includegraphics[width=1.0\textwidth]{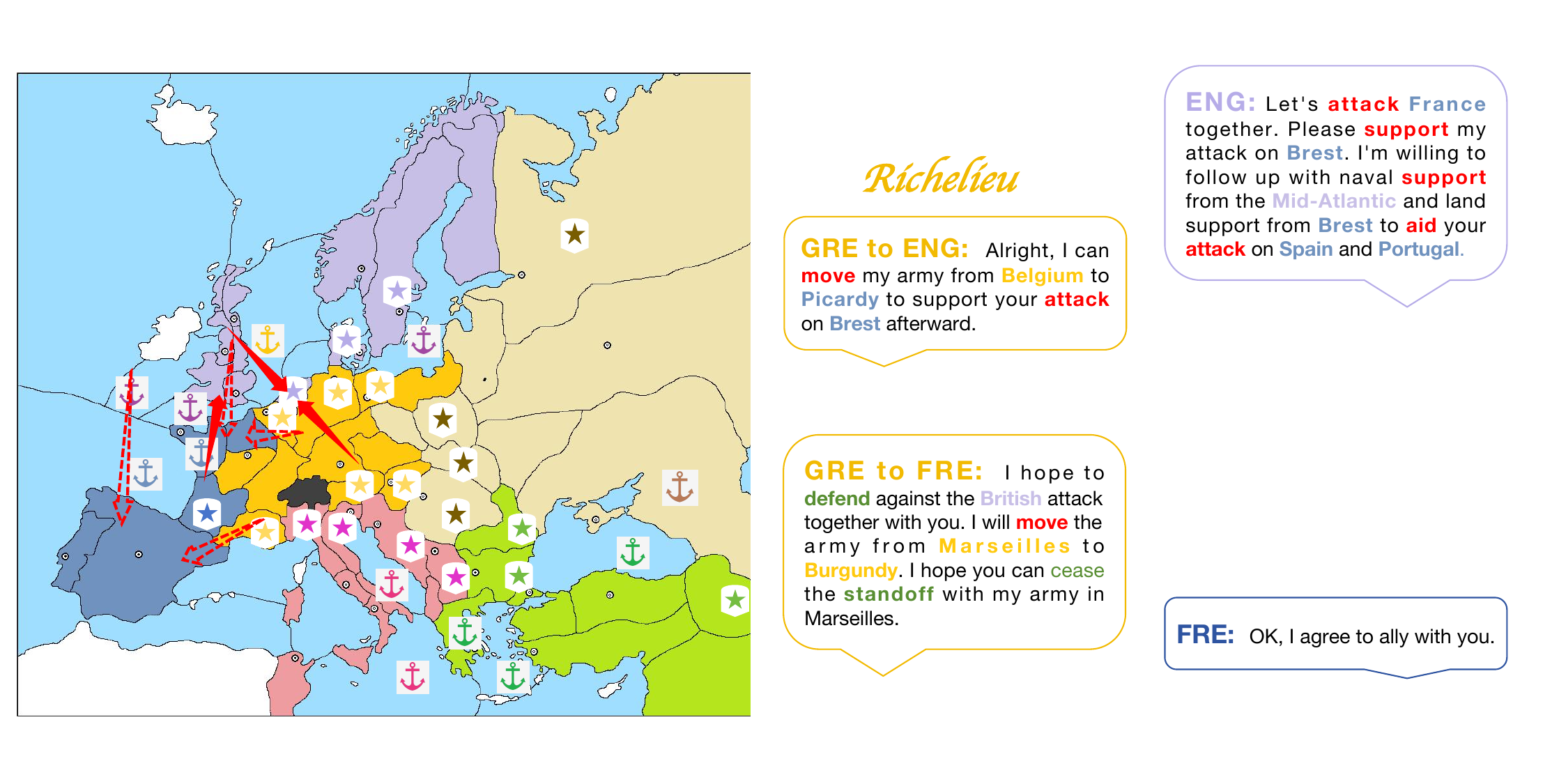}  
    \caption{An example case of avoiding being deceived by other countries during negotiations. }
    \label{case2}
\end{figure}

As shown in Figure ~\ref{case2}, Richelieu controls Germany. During the negotiation phase, England proposed a ceasefire to Germany and invited Germany to form an alliance to attack France jointly. England hoped to cease the war with Germany in Holland and Belgium. Subsequently, German units supported England in attacking Brest, and then England utilized its fleets to assist Germany in attacking Spain and Portugal. Richelieu suspected that England was deceiving Germany, as England was likely to attack territories in the north such as Belgium and Berlin after German units were redirected to support Brest. Therefore, we pretended to accept England's alliance proposal during the negotiation process. However, at the same time, we sought out France and expressed our willingness to cease hostilities, allowing France to focus entirely on defending against England's attacks. 
In the action phase, England's actions confirmed Richelieu's suspicions. England attacked Belgium from Holland, but because Richelieu didn't move units in Belgium, England's attack failed.

\section{More application}
Our modules cover most of the challenges in multi-agent interactions, e.g., economic games, and daily interactions. 

To prove that our framework is capable of applying to most social interaction tasks, we further adopt our framework to a werewolf game. The results demonstrate our reasoning framework achieves comparable results to the other methods. To be specific, in the experiment, we let our agent play as a werewolf in a seven-player game, where there are two werewolves, one witch, one seer, one guard, and two villagers. The experimental results show that the win rate of our agent is 59.2\%, even without applying the self-play game in the current version. For comparison, the strongest specifically designed LLM-based agent achieved ~65\% win rate \citet{xu2023exploring}.
This proves that our model can be applied in more scenarios and achieve results comparable to those of specially designed models.

\section{Ethical Consideration}

The method proposed in this work has the potential for positive uses like enabling AI agents to emerge in cooperation via negotiation or avoiding being fooled by fake promises (or helping humans do so). However, negative cases can also arise if the technique is used for possible fraud activities. Fortunately, there is research \citet{bakhtin2019real}\citet{zellers2019defending} dealing with such scenarios. And we also urge for more research efforts in this field to foster safe applications of similar technologies.

\newpage
\section*{NeurIPS Paper Checklist}

The checklist is designed to encourage best practices for responsible machine learning research, addressing issues of reproducibility, transparency, research ethics, and societal impact. Do not remove the checklist: {\bf The papers not including the checklist will be desk rejected.} The checklist should follow the references and follow the (optional) supplemental material.  The checklist does NOT count toward the page limit. 

Please read the checklist guidelines carefully for information on how to answer these questions. For each question in the checklist:
\begin{itemize}
    \item You should answer \answerYes{}, \answerNo{}, or \answerNA{}.
    \item \answerNA{} means either that the question is Not Applicable for that particular paper or the relevant information is Not Available.
    \item Please provide a short (1–2 sentence) justification right after your answer (even for NA). 
\end{itemize}

{\bf The checklist answers are an integral part of your paper submission.} They are visible to the reviewers, area chairs, senior area chairs, and ethics reviewers. You will be asked to also include it (after eventual revisions) with the final version of your paper, and its final version will be published with the paper.

The reviewers of your paper will be asked to use the checklist as one of the factors in their evaluation. While "\answerYes{}" is generally preferable to "\answerNo{}", it is perfectly acceptable to answer "\answerNo{}" provided a proper justification is given (e.g., "error bars are not reported because it would be too computationally expensive" or "we were unable to find the license for the dataset we used"). In general, answering "\answerNo{}" or "\answerNA{}" is not grounds for rejection. While the questions are phrased in a binary way, we acknowledge that the true answer is often more nuanced, so please just use your best judgment and write a justification to elaborate. All supporting evidence can appear either in the main paper or the supplemental material, provided in the appendix. If you answer \answerYes{} to a question, in the justification please point to the section(s) where related material for the question can be found.

IMPORTANT, please:
\begin{itemize}
    \item {\bf Delete this instruction block, but keep the section heading ``NeurIPS paper checklist"},
    \item  {\bf Keep the checklist subsection headings, questions/answers, and guidelines below.}
    \item {\bf Do not modify the questions and only use the provided macros for your answers}.
\end{itemize}


\begin{enumerate}

\item {\bf Claims}
    \item[] Question: Do the main claims made in the abstract and introduction accurately reflect the paper's contributions and scope?
    \item[] Answer: \answerYes{} 
    \item[] Justification: The contributions and scope has been fully covered by the abstract and introduction sections.
    \item[] Guidelines:
    \begin{itemize}
        \item The answer NA means that the abstract and introduction do not include the claims made in the paper.
        \item The abstract and/or introduction should clearly state the claims made, including the contributions made in the paper and important assumptions and limitations. A No or NA answer to this question will not be perceived well by the reviewers. 
        \item The claims made should match theoretical and experimental results, and reflect how much the results can be expected to generalize to other settings. 
        \item It is fine to include aspirational goals as motivation as long as it is clear that these goals are not attained by the paper. 
    \end{itemize}

\item {\bf Limitations}
    \item[] Question: Does the paper discuss the limitations of the work performed by the authors?
    \item[] Answer: \answerYes{} 
    \item[] Justification: The paper discussed the limitations of the work performed by the authors in the section "Limitation and Future Work".
    \item[] Guidelines:
    \begin{itemize}
        \item The answer NA means that the paper has no limitation while the answer No means that the paper has limitations, but those are not discussed in the paper. 
        \item The authors are encouraged to create a separate "Limitations" section in their paper.
        \item The paper should point out any strong assumptions and how robust the results are to violations of these assumptions (e.g., independence assumptions, noiseless settings, model well-specification, asymptotic approximations only holding locally). The authors should reflect on how these assumptions might be violated in practice and what the implications would be.
        \item The authors should reflect on the scope of the claims made, e.g., if the approach was only tested on a few datasets or with a few runs. In general, empirical results often depend on implicit assumptions, which should be articulated.
        \item The authors should reflect on the factors that influence the performance of the approach. For example, a facial recognition algorithm may perform poorly when the image resolution is low or images are taken in low lighting. Or a speech-to-text system might not be used reliably to provide closed captions for online lectures because it fails to handle technical jargon.
        \item The authors should discuss the computational efficiency of the proposed algorithms and how they scale with dataset size.
        \item If applicable, the authors should discuss possible limitations of their approach to address problems of privacy and fairness.
        \item While the authors might fear that complete honesty about limitations might be used by reviewers as grounds for rejection, a worse outcome might be that reviewers discover limitations that aren't acknowledged in the paper. The authors should use their best judgment and recognize that individual actions in favor of transparency play an important role in developing norms that preserve the integrity of the community. Reviewers will be specifically instructed to not penalize honesty concerning limitations.
    \end{itemize}

\item {\bf Theory Assumptions and Proofs}
    \item[] Question: For each theoretical result, does the paper provide the full set of assumptions and a complete (and correct) proof?
    \item[] Answer: \answerYes{} 
    \item[] Justification: The paper provides the full set of assumptions and a complete (and correct) proof for each theoretical result in the "Method" and "Experiment" sections.
    \item[] Guidelines:
    \begin{itemize}
        \item The answer NA means that the paper does not include theoretical results. 
        \item All the theorems, formulas, and proofs in the paper should be numbered and cross-referenced.
        \item All assumptions should be clearly stated or referenced in the statement of any theorems.
        \item The proofs can either appear in the main paper or the supplemental material, but if they appear in the supplemental material, the authors are encouraged to provide a short proof sketch to provide intuition. 
        \item Inversely, any informal proof provided in the core of the paper should be complemented by formal proofs provided in the appendix or supplemental material.
        \item Theorems and Lemmas that the proof relies upon should be properly referenced. 
    \end{itemize}

    \item {\bf Experimental Result Reproducibility}
    \item[] Question: Does the paper fully disclose all the information needed to reproduce the main experimental results of the paper to the extent that it affects the main claims and/or conclusions of the paper (regardless of whether the code and data are provided or not)?
    \item[] Answer: \answerYes{} 
    \item[] Justification: The paper fully discloses all the information needed to reproduce the main experimental results in the main text section "Experiment" and appendix section "Implementation Details".
    \item[] Guidelines:
    \begin{itemize}
        \item The answer NA means that the paper does not include experiments.
        \item If the paper includes experiments, a No answer to this question will not be perceived well by the reviewers: Making the paper reproducible is important, regardless of whether the code and data are provided or not.
        \item If the contribution is a dataset and/or model, the authors should describe the steps taken to make their results reproducible or verifiable. 
        \item Depending on the contribution, reproducibility can be accomplished in various ways. For example, if the contribution is a novel architecture, describing the architecture fully might suffice, or if the contribution is a specific model and empirical evaluation, it may be necessary to either make it possible for others to replicate the model with the same dataset, or provide access to the model. In general. releasing code and data is often one good way to accomplish this, but reproducibility can also be provided via detailed instructions for how to replicate the results, access to a hosted model (e.g., in the case of a large language model), releasing of a model checkpoint, or other means that are appropriate to the research performed.
        \item While NeurIPS does not require releasing code, the conference does require all submissions to provide some reasonable avenue for reproducibility, which may depend on the nature of the contribution. For example
        \begin{enumerate}
            \item If the contribution is primarily a new algorithm, the paper should make it clear how to reproduce that algorithm.
            \item If the contribution is primarily a new model architecture, the paper should describe the architecture clearly and fully.
            \item If the contribution is a new model (e.g., a large language model), then there should either be a way to access this model for reproducing the results or a way to reproduce the model (e.g., with an open-source dataset or instructions for how to construct the dataset).
            \item We recognize that reproducibility may be tricky in some cases, in which case authors are welcome to describe the particular way they provide for reproducibility. In the case of closed-source models, it may be that access to the model is limited in some way (e.g., to registered users), but it should be possible for other researchers to have some path to reproducing or verifying the results.
        \end{enumerate}
    \end{itemize}

\item {\bf Open access to data and code}
    \item[] Question: Does the paper provide open access to the data and code, with sufficient instructions to faithfully reproduce the main experimental results, as described in supplemental material?
    \item[] Answer: \answerYes{} 
    \item[] Justification: The paper provides open access to the data and code in the section "Experiment".
    \item[] Guidelines:
    \begin{itemize}
        \item The answer NA means that paper does not include experiments requiring code.
        \item Please see the NeurIPS code and data submission guidelines (\url{https://nips.cc/public/guides/CodeSubmissionPolicy}) for more details.
        \item While we encourage the release of code and data, we understand that this might not be possible, so “No” is an acceptable answer. Papers cannot be rejected simply for not including code, unless this is central to the contribution (e.g., for a new open-source benchmark).
        \item The instructions should contain the exact command and environment needed to run to reproduce the results. See the NeurIPS code and data submission guidelines (\url{https://nips.cc/public/guides/CodeSubmissionPolicy}) for more details.
        \item The authors should provide instructions on data access and preparation, including how to access the raw data, preprocessed data, intermediate data, and generated data, etc.
        \item The authors should provide scripts to reproduce all experimental results for the new proposed method and baselines. If only a subset of experiments are reproducible, they should state which ones are omitted from the script and why.
        \item At submission time, to preserve anonymity, the authors should release anonymized versions (if applicable).
        \item Providing as much information as possible in supplemental material (appended to the paper) is recommended, but including URLs to data and code is permitted.
    \end{itemize}

\item {\bf Experimental Setting/Details}
    \item[] Question: Does the paper specify all the training and test details (e.g., data splits, hyperparameters, how they were chosen, type of optimizer, etc.) necessary to understand the results?
    \item[] Answer: \answerYes{} 
    \item[] Justification: The paper specify all the training and test details to train the model in the "Experiment" section and "Implementation Details" section of appendix.
    \item[] Guidelines:
    \begin{itemize}
        \item The answer NA means that the paper does not include experiments.
        \item The experimental setting should be presented in the core of the paper to a level of detail that is necessary to appreciate the results and make sense of them.
        \item The full details can be provided either with the code, in appendix, or as supplemental material.
    \end{itemize}

\item {\bf Experiment Statistical Significance}
    \item[] Question: Does the paper report error bars suitably and correctly defined or other appropriate information about the statistical significance of the experiments?
    \item[] Answer: \answerYes{} 
    \item[] Justification: The paper reports error bars suitably and correctly defined or other appropriate information about the statistical significance of the experiment in the "Experiment" section.
    \item[] Guidelines:
    \begin{itemize}
        \item The answer NA means that the paper does not include experiments.
        \item The authors should answer "Yes" if the results are accompanied by error bars, confidence intervals, or statistical significance tests, at least for the experiments that support the main claims of the paper.
        \item The factors of variability that the error bars are capturing should be clearly stated (for example, train/test split, initialization, random drawing of some parameter, or overall run with given experimental conditions).
        \item The method for calculating the error bars should be explained (closed form formula, call to a library function, bootstrap, etc.)
        \item The assumptions made should be given (e.g., Normally distributed errors).
        \item It should be clear whether the error bar is the standard deviation or the standard error of the mean.
        \item It is OK to report 1-sigma error bars, but one should state it. The authors should preferably report a 2-sigma error bar than state that they have a 96\% CI, if the hypothesis of Normality of errors is not verified.
        \item For asymmetric distributions, the authors should be careful not to show in tables or figures symmetric error bars that would yield results that are out of range (e.g. negative error rates).
        \item If error bars are reported in tables or plots, The authors should explain in the text how they were calculated and reference the corresponding figures or tables in the text.
    \end{itemize}

\item {\bf Experiments Compute Resources}
    \item[] Question: For each experiment, does the paper provide sufficient information on the computer resources (type of compute workers, memory, time of execution) needed to reproduce the experiments?
    \item[] Answer: \answerYes{} 
    \item[] Justification: For each experiment, the paper provide sufficient information on the computer resources.
    \item[] Guidelines:
    \begin{itemize}
        \item The answer NA means that the paper does not include experiments.
        \item The paper should indicate the type of compute worker CPU or GPU, internal cluster, or cloud provider, including relevant memory and storage.
        \item The paper should provide the amount of compute required for each of the individual experimental runs as well as estimate the total compute. 
        \item The paper should disclose whether the full research project required more compute than the experiments reported in the paper (e.g., preliminary or failed experiments that didn't make it into the paper). 
    \end{itemize}
    
\item {\bf Code Of Ethics}
    \item[] Question: Does the research conducted in the paper conform, in every respect, with the NeurIPS Code of Ethics \url{https://neurips.cc/public/EthicsGuidelines}?
    \item[] Answer: \answerYes{} 
    \item[] Justification: The research conducted in the paper conform, in every respect, comply with the NeurIPS Code of Ethics.
    \item[] Guidelines:
    \begin{itemize}
        \item The answer NA means that the authors have not reviewed the NeurIPS Code of Ethics.
        \item If the authors answer No, they should explain the special circumstances that require a deviation from the Code of Ethics.
        \item The authors should make sure to preserve anonymity (e.g., if there is a special consideration due to laws or regulations in their jurisdiction).
    \end{itemize}

\item {\bf Broader Impacts}
    \item[] Question: Does the paper discuss both potential positive societal impacts and negative societal impacts of the work performed?
    \item[] Answer: \answerYes{} 
    \item[] Justification: The paper discuss both potential positive societal impacts and negative societal impacts of the work performed in Appendix section "Ethical Consideration".
    \item[] Guidelines:
    \begin{itemize}
        \item The answer NA means that there is no societal impact of the work performed.
        \item If the authors answer NA or No, they should explain why their work has no societal impact or why the paper does not address societal impact.
        \item Examples of negative societal impacts include potential malicious or unintended uses (e.g., disinformation, generating fake profiles, surveillance), fairness considerations (e.g., deployment of technologies that could make decisions that unfairly impact specific groups), privacy considerations, and security considerations.
        \item The conference expects that many papers will be foundational research and not tied to particular applications, let alone deployments. However, if there is a direct path to any negative applications, the authors should point it out. For example, it is legitimate to point out that an improvement in the quality of generative models could be used to generate deepfakes for disinformation. On the other hand, it is not needed to point out that a generic algorithm for optimizing neural networks could enable people to train models that generate Deepfakes faster.
        \item The authors should consider possible harms that could arise when the technology is being used as intended and functioning correctly, harms that could arise when the technology is being used as intended but gives incorrect results, and harms following from (intentional or unintentional) misuse of the technology.
        \item If there are negative societal impacts, the authors could also discuss possible mitigation strategies (e.g., gated release of models, providing defenses in addition to attacks, mechanisms for monitoring misuse, mechanisms to monitor how a system learns from feedback over time, improving the efficiency and accessibility of ML).
    \end{itemize}
    
\item {\bf Safeguards}
    \item[] Question: Does the paper describe safeguards that have been put in place for responsible release of data or models that have a high risk for misuse (e.g., pretrained language models, image generators, or scraped datasets)?
    \item[] Answer: \answerYes{} 
    \item[] Justification: The paper describe safeguards that have been put in place for responsible release of data or models that have a high risk for misuse in Appendix section "Ethical Consideration".
    \item[] Guidelines:
    \begin{itemize}
        \item The answer NA means that the paper poses no such risks.
        \item Released models that have a high risk for misuse or dual-use should be released with necessary safeguards to allow for controlled use of the model, for example by requiring that users adhere to usage guidelines or restrictions to access the model or implementing safety filters. 
        \item Datasets that have been scraped from the Internet could pose safety risks. The authors should describe how they avoided releasing unsafe images.
        \item We recognize that providing effective safeguards is challenging, and many papers do not require this, but we encourage authors to take this into account and make a best faith effort.
    \end{itemize}

\item {\bf Licenses for existing assets}
    \item[] Question: Are the creators or original owners of assets (e.g., code, data, models), used in the paper, properly credited and are the license and terms of use explicitly mentioned and properly respected?
    \item[] Answer: \answerYes{} 
    \item[] Justification: the creators or original owners of assets (e.g., code, data, models), used in the paper, properly credited and are the license and terms of use explicitly mentioned and properly respected
    \item[] Guidelines:
    \begin{itemize}
        \item The answer NA means that the paper does not use existing assets.
        \item The authors should cite the original paper that produced the code package or dataset.
        \item The authors should state which version of the asset is used and, if possible, include a URL.
        \item The name of the license (e.g., CC-BY 4.0) should be included for each asset.
        \item For scraped data from a particular source (e.g., website), the copyright and terms of service of that source should be provided.
        \item If assets are released, the license, copyright information, and terms of use in the package should be provided. For popular datasets, \url{paperswithcode.com/datasets} has curated licenses for some datasets. Their licensing guide can help determine the license of a dataset.
        \item For existing datasets that are re-packaged, both the original license and the license of the derived asset (if it has changed) should be provided.
        \item If this information is not available online, the authors are encouraged to reach out to the asset's creators.
    \end{itemize}

\item {\bf New Assets}
    \item[] Question: Are new assets introduced in the paper well documented and is the documentation provided alongside the assets?
    \item[] Answer: \answerYes{} 
    \item[] Justification: new assets introduced in the paper are well documented and the documentation is provided in section "Experiment".
    \item[] Guidelines:
    \begin{itemize}
        \item The answer NA means that the paper does not release new assets.
        \item Researchers should communicate the details of the dataset/code/model as part of their submissions via structured templates. This includes details about training, license, limitations, etc. 
        \item The paper should discuss whether and how consent was obtained from people whose asset is used.
        \item At submission time, remember to anonymize your assets (if applicable). You can either create an anonymized URL or include an anonymized zip file.
    \end{itemize}

\item {\bf Crowdsourcing and Research with Human Subjects}
    \item[] Question: For crowdsourcing experiments and research with human subjects, does the paper include the full text of instructions given to participants and screenshots, if applicable, as well as details about compensation (if any)? 
    \item[] Answer: \answerYes{} 
    \item[] Justification: our work does not involve crowdsourcing or research with human subjects.
    \item[] Guidelines:
    \begin{itemize}
        \item The answer NA means that the paper does not involve crowdsourcing nor research with human subjects.
        \item Including this information in the supplemental material is fine, but if the main contribution of the paper involves human subjects, then as much detail as possible should be included in the main paper. 
        \item According to the NeurIPS Code of Ethics, workers involved in data collection, curation, or other labor should be paid at least the minimum wage in the country of the data collector. 
    \end{itemize}

\item {\bf Institutional Review Board (IRB) Approvals or Equivalent for Research with Human Subjects}
    \item[] Question: Does the paper describe potential risks incurred by study participants, whether such risks were disclosed to the subjects, and whether Institutional Review Board (IRB) approvals (or an equivalent approval/review based on the requirements of your country or institution) were obtained?
    \item[] Answer: \answerYes{} 
    \item[] Justification: our work does not involve research with human subjects.
    \item[] Guidelines:
    \begin{itemize}
        \item The answer NA means that the paper does not involve crowdsourcing nor research with human subjects.
        \item Depending on the country in which research is conducted, IRB approval (or equivalent) may be required for any human subjects research. If you obtained IRB approval, you should clearly state this in the paper. 
        \item We recognize that the procedures for this may vary significantly between institutions and locations, and we expect authors to adhere to the NeurIPS Code of Ethics and the guidelines for their institution. 
        \item For initial submissions, do not include any information that would break anonymity (if applicable), such as the institution conducting the review.
    \end{itemize}

\end{enumerate}

\end{document}